\pdfoutput=1

\documentclass[journal]{IEEEtran}

\usepackage[english]{babel}
\usepackage{fancyhdr,amsmath,amsthm,amssymb,url,array,textcomp,listings,xcolor,colortbl,float,gensymb,longtable,supertabular,multicol,placeins,tabularx}
\usepackage{booktabs}
\usepackage{multirow}
\usepackage{pbox,nth}
\usepackage[flushleft]{threeparttable}

\usepackage{xcolor}
\usepackage[printwatermark]{xwatermark}
\newwatermark*[allpages,color=red,angle=0,scale=1,xpos=0,ypos=373pt,fontsize=6pt]{
The content of this paper was published in Neurocomputing, 2017. This ArXiv version is from before the peer review. Please, cite the following paper:
\\Ziga Emersic, Vitomir Struc, Peter Peer: "Ear Recognition: More Than a Survey", Neurocomputing, 2017
}

\usepackage{graphicx,subfig}

\usepackage[utf8x]{inputenc}

\usepackage[T1]{fontenc}
\usepackage{lmodern}
\input{glyphtounicode}
\begin{document}




\title{Ear Recognition: More Than a Survey}



\author{Žiga~Emeršič, Vitomir~Štruc, Peter Peer
\thanks{Ž. Emeršič and Peter Peer are with the Faculty of Computer and Information Science, University of Ljubljana, e-mail: \{ziga.emersic, peter.peer\}@fri.uni-lj.si.}
\thanks{V. Štruc is with the Faculty of Electrical Engineering, University of Ljubljana, e-mail: vitomir.struc@fe.uni-lj.si.}
\thanks{\textcolor{red}{Accepted for publication in Neurocomputing on August 24, 2016.}}
}

\maketitle

\begin{abstract}
 Automatic identity recognition from ear images represents an active field of research within the biometric community. The ability to capture ear images from a distance and in a covert manner makes the technology an appealing choice for surveillance and security applications as well as other application domains. Significant contributions have been made in the field over recent years, but open research problems still remain and hinder a wider (commercial) deployment of the technology. This paper presents an overview of the field of automatic ear recognition (from 2D images) and focuses specifically on the most recent, descriptor-based methods proposed in this area. Open challenges are discussed and potential research directions are outlined with the goal of providing the reader with a point of reference for issues worth examining in the future. In addition to a comprehensive review on ear recognition technology, the paper also introduces a new, fully unconstrained dataset of ear images gathered from the web and a toolbox implementing several state-of-the-art techniques for ear recognition. The dataset and toolbox are meant to address some of the open issues in the field and are made publicly available to the research community.
\end{abstract}

\begin{IEEEkeywords}
biometry, dataset, in-the-wild, unconstrained image, descriptor-based method, open-source toolbox, ear recognition
\end{IEEEkeywords}


\newcommand{\dname}{AWE }
\definecolor{earOrange}{rgb}{1,0.5,0}
\definecolor{earGreen}{rgb}{0.05,0.64,0.05}

\newif\ifcolordiff
\colordifffalse

\ifcolordiff
	\newcommand\del[1]{{\color{red}#1}} 
	\newcommand\add[1]{{\color{blue}#1}} 
	\newcommand\mvf[1]{{\color{earOrange}#1}} 
	\newcommand\mvt[1]{{\color{earGreen}#1}}
\else
	\newcommand\del[1]{{\iffalse#1\fi}}
	\newcommand\add[1]{{\color{black}#1}}
	\newcommand\mvf[1]{{\iffalse#1\fi}}
	\newcommand\mvt[1]{{\color{black}#1}}
\fi


\section{Introduction}
\label{section:introduction}


\begin{figure*}
	\centering
	\includegraphics[width=0.9\textwidth]{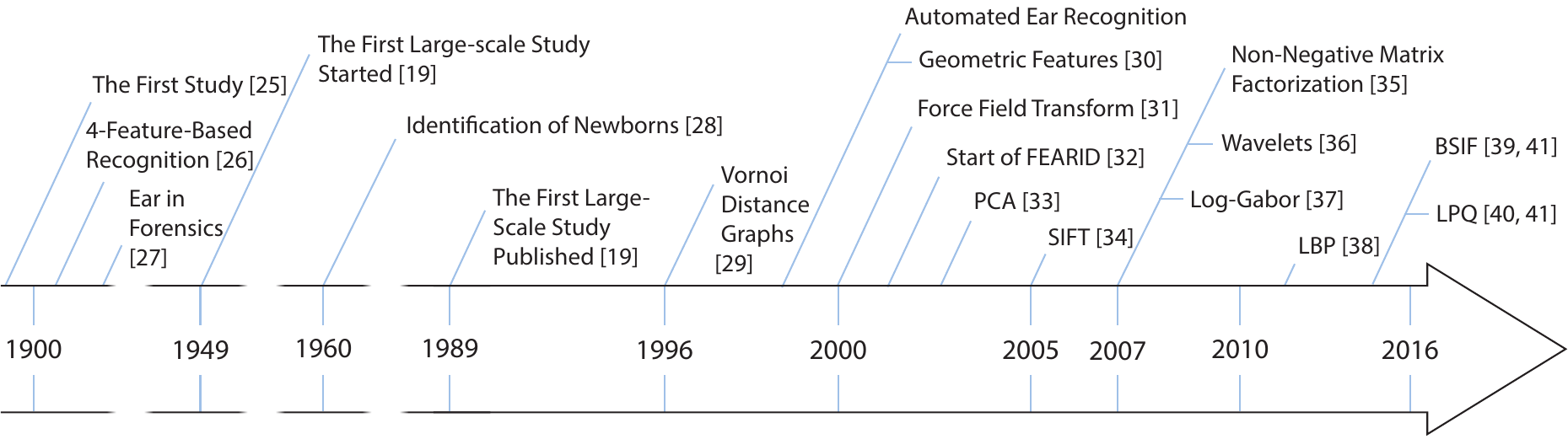}
	\caption{Development of ear recognition approaches through time. S\del{ome s}elected milestones are presented.}
	\label{figure:approaches_timeline}
\end{figure*}


\mvt{Ear images used in automatic ear recognition systems can typically be extracted from profile head shots or video footage. The acquisition procedure is \textit{contactless} and \textit{nonintrusive} and also \textit{does not depend on the cooperativeness} of the person one is trying to recognize. In this regard ear recognition technology shares similarities with other image-based biometric modalities. Another appealing property of ear biometrics is its \textit{distinctiveness}~\cite{Jain2011}. Recent studies even empirically validated existing conjectures that certain features of the ear are distinct for identical twins~\cite{Sim2012ICPR}. This fact has significant implications for security related applications and puts ear images on par with epigenetic biometric modalities, such as the iris. Ear images can also serve as \textit{supplements} for other biometric modalities in automatic recognition systems and provide identity cues when other information is unreliable or even unavailable. In surveillance applications, for example, where face recognition technology may struggle with profile faces, the ear can serve as a source of information on the identity of people in the surveillance footage. The importance and potential value of ear recognition technology for multi-modal biometric systems is also evidenced by the number of research studies on this topic, e.g.~\cite{MM2007,MM2008,MM2008b,MM2009,MM2009b}.

Today, ear recognition represents an active research area, for which new techniques are developed on a regular basis and several datasets needed for training and testing of the technology are publicly available, e.g.,~\cite{Frejlichowski2010,Kumar2012}. Nevertheless, despite the research efforts directed at ear biometrics, to the best of our knowledge, there is only one commercial system currently available on the marked that exploits ear biometrics for recognition, i.e., the Helix from Descartes Biometrics~\cite{Helix2015}. We conjecture that the limited availability of the commercial ear recognition technology is a consequence of the open challenges that by today have still not been appropriately addressed. This paper is an attempt to meet some of these challenges and provide the community with a point of reference as well as with new research tools that can be used to further advance the field.}

\del{The first observations regarding the uniqueness of the outer ear and its capability for identity recognition  date back to the end of the 19th century and are attributed to the French police officer Bertillon~\cite{Bertillon1896}, who developed an anthropometric system to identify criminals.  Similar observations about the potential of ears for establishing identity were also published a few years later by a doctor from Prague, named Imhofer~\cite{Imhofer1906}, who succeeded to distinguish between ears of 500 individuals based on only four distinct ear features. 
In 1909 Locard~\cite{Locard1909} discussed the importance of the ear in forensic investigations. A milestone for ear recognition marked the study of Fields et al.~\cite{Fields1960} from 1960, where the shape of the outer ear was used to distinguish between 206 newborns. While finger and palm-prints turned out to be ineffective for this task, the ears were found to be unique for every newborn included in the study.}

\mvf{One of the biggest contributions to the field of ear recognition was made by Iannarelli in 1989~\cite{Iannarelli1989}, when he published a four decades long study on the potential of ear recognition. Iannarelli's seminal work included more than $10,000$ ears and addressed various aspects of recognition, such as ear similarity of siblings, twins and triplets, relations between the appearance of the ears of parents and children as well as racial variations of ear appearance~\cite{Purkait2015}.

The 1990s marked the beginning of automatic ear recognition. Various methods were developed during this time and introduced in the literature. In 1996, for example, Burge and Burger~\cite{Burge1996} used adjacency graphs computed from Voronoi diagrams of the ears curve segments for ear description and in 1999 Moreno et al.~\cite{Moreno1999} presented the first fully automated ear recognition procedure exploiting geometric characteristics of the ear and a compression network. In 2000 Hurley et al.~\cite{Hurley2000} described an approach for ear recognition that relied on the Force Field Transform, which proved highly successful for this task. A year later, in 2001, the Forensic Ear Identification Project (FEARID) project was launched, marking the first large-scale project in the field of ear recognition~\cite{Alberink2007}.

With the beginning of the new millennium, automatic ear recognition techniques started to gain traction with the biometric community and new techniques were introduced more frequently. In 2002 Victor et al.~\cite{Victor2002} applied principal component analysis (PCA) on ear images and reported promising results. In 2005 the Scale Invariant Feature Transform (SIFT)~\cite{Hurley2005} was used for the first time with ear images, raising the bar for the performance of existing recognition techniques. In 2006 a method based on non-negative matrix factorization (NMF) was developed by Yuan et al.~\cite{Yuan2006} and applied to occluded and non-occluded ear images with competitive results. In 2007 a method based on the 2D wavelet transform was introduced by Nosrati et al.~\cite{Nosrati2007}, followed by a technique based on log-Gabor wavelets in the same year~\cite{Kumar2007}. More recently, in 2011, local binary patterns (LBP) were used for ear-image description in~\cite{Zhi-Qin2011}, while later binarized statistical image features (BSIF) and local phase quantization (LPQ) features also proved successful for this task~\cite{Benzaoui2014, Pflug2014,Pflug2014a}. A graphical representation of the main milestones\footnote{In the opinion of the authors.} in the development of ear recognition technology (briefly discussed above) is shown in Fig.~\ref{figure:approaches_timeline}.

Today, ear recognition represents an active research area, for which new techniques are developed on a regular basis and several datasets needed for training and testing of the technology are publicly available, e.g.,~\cite{Frejlichowski2010,Kumar2012}. Nevertheless, despite the research efforts directed at ear biometrics, to the best of our knowledge, there is only one commercial system currently available on the marked that exploits ear biometrics for recognition, i.e., the Helix from Descartes Biometrics~\cite{Helix2015}. We conjecture that the limited availability of commercial ear recognition technology is a consequence of the open challenges that by today have still not been appropriately addressed. This paper is an attempt to meet some of these challenges and provide the community with a point of reference as well as with new research tools that can be used to further advance the field.}

\subsection{Contributions and Paper Organization}

Prior surveys related to ear recognition, such as~\cite{Purkait2015,Islam2008Survey,Choras2007,PoonMoon2004,NixonSurvey2013,Pflug2012}, provide well written and well structured reviews of the field. In this paper we contribute to these surveys by discussing recent 2D ear recognition techniques proposed until the end of 2015. We pay special attention to descriptor-based approaches that are currently considered state-of-the-art in 2D ear recognition.
We present comparative experiments with the new dataset and toolbox to establish an independent ranking of the state-of-the-art techniques and show that there is significant room for improvement and that ear recognition is far from being solved.

We make the following contributions in this paper:
\begin{itemize}
\item \add{\textbf{Survey:}} We present a comprehensive survey on ear recognition, which is meant to provide researchers in this field with a recent and up-to-date overview of the state-of-technology. We introduce a taxonomy of the existing 2D ear recognition approaches, discuss the characteristics of the technology and review the existing state-of-the-art. 
    Most importantly, we elaborate on the open problems and challenges faced by the technology.
\item \add{\textbf{Dataset:}} We make a new dataset of ear images available to the research community. The dataset, named Annotated Web Ears (AWE), contains images collected from the web and is to the best of our knowledge the first dataset for ear recognition gathered ``in the wild''. The images of the \dname dataset contain a high degree of variability and present a challenging problem to the existing technology, as shown in \del{Fig.~\ref{figure:first_graphs}}\add{the experimental section}. \del{Here, the rank-1 recognition rates and verification rates at 1\% false acceptance rate are presented for a few popular ear datasets and the best performing method from our experiments for each dataset (see Section~\ref{section:experiments}).}
\item \add{\textbf{Toolbox:}} We introduce an open source (Matlab) toolbox, i.e., the \dname toolbox,  for research in ear recognition. The toolbox implements a number of state-of-the-art feature extraction techniques as well as other important steps in the processing pipeline of ear recognition systems. It contains tools for generating performance metrics and graphs and allows for transparent and reproducible research in ear recognition. The toolbox is available from: \url{http://awe.fri.uni-lj.si}.
\item \add{\textbf{Reproducible evaluation:}} We conduct a comparative evaluation of several state-of-the-art methods on a number of popular ear datasets using consistent experimental protocols, which enables direct comparisons of the state-of-the-art in ear recognition. All experiments are conducted with our \dname toolbox making all presented results reproducible.
\end{itemize}
The rest of the paper is structured as follows. In Section~\ref{section:high-level} we present the background and basic terminology related to ear recognition.
In Section~\ref{section:data} existing ear datasets are discussed and compared on the basis of some common criteria of interest. Section~\ref{Section:DatasetAndToolbox} introduces our new \dname dataset and the accompanying \dname toolbox. Comparative experiments and results with the new toolbox are presented in Section~\ref{section:experiments}. In Section~\ref{section:research} open problems and promising future research directions are examined. The paper concludes with some final comments in Section~\ref{section:conclusion}.

\section{Ear Recognition Essentials}
\label{section:high-level}

\del{In this section we introduce the basic terminology related to ear recognition and present a brief overview of existing approaches used in the field. We commence the section with a description of the structure of the ear and its potential for identity recognition. Next, we survey existing approaches and discuss our taxonomy that will be used throughout the paper. Finally, we elaborate on the characteristics of ears as a biometric modality.}

\subsection{Ear Structure}

\begin{figure}[]
	\centering
	\includegraphics[width=0.65\columnwidth]{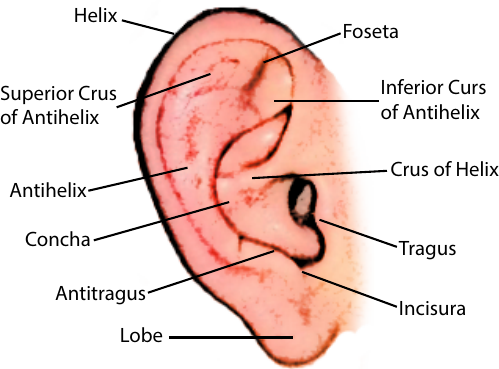}
	\caption{Ear structure.}
	\label{figure:ear_structure}
\end{figure}

The human ear develops early during pregnancy and is already fully formed by the time of birth. Due to its role as the human hearing organ, the ear has a characteristic structure that is (for the most part) shared across the population. The appearance of the outer ear is defined by the shapes of the tragus, the antitragus, the helix, the antihelix, the incisura, the lope and other important structural parts as shown in Fig.~\ref{figure:ear_structure}. These anatomical cartilage formations differ in shape, appearance and relative positions from person to person and can, therefore, be exploited for identity recognition.

In general, the left and right ears of a person are similar to such an extent that makes matching the right ear to the left ear (and vice versa) with automatic techniques perform significantly better than chance. Yan and Bowyer~\cite{Yan2005Symm}, for example, reported a recognition performance of around 90\% in cross-ear matching experiments. They observed that for most people the left and right ears are at least close to bilateral symmetric, though the shape of the two ears is different for some~\cite{Yan2005Symm}. Similar findings were also reported by Abaza and Ross in~\cite{SymAbaza2010}.

The literature suggests that the size of the ear changes through time~\cite{Iannarelli1989,NixonSurvey2013,Pflug2012,Sforza2009}. Longitudinal studies from India~\cite{Purkait2007} and Europe~\cite{Sforza2009,Gualdi1998,Ferrario1999} have found that the length of the ear increases significantly with age for men and women, while the width remains relatively constant. How ear growth affects the performance of automatic recognition system is currently still an open research question. The main problem here is the lack of appropriate datasets captured over a long enough period of time that could help provide final and conclusive answers. Some initial studies appeared recently on this topic, but only featured images captured less than a year apart~\cite{Ibrahim2011IJCB}.

\add{\subsection{Chronological Development of Ear Recognition}}

\mvt{The chronological development of ear recognition techniques can be divided into a \textit{manual (pre-automatic)} and \textit{automatic era}. During the pre-automatic era several studies and empirical observations were published pointing to the potential of ears for identity recognition~\cite{Bertillon1896,Imhofer1906,Locard1909,Fields1960}. One of the biggest contributions to the field during this era was made by Iannarelli in 1989~\cite{Iannarelli1989}, when he published a long-term study on the potential of ear recognition. Iannarelli's seminal work included more than $10,000$ ears and addressed various aspects of recognition, such as ear similarity of siblings, twins and triplets, relations between the appearance of the ears of parents and children as well as racial variations of ear appearance~\cite{Purkait2015}.

The 1990s marked the beginning of automatic ear recognition. Various methods were developed during this time and were introduced in the literature. In 1996, for example, Burge and Burger~\cite{Burge1996} used adjacency graphs computed from Voronoi diagrams of the ears curve segments for ear description and in 1999 Moreno et al.~\cite{Moreno1999} presented the first fully automated ear recognition procedure exploiting geometric characteristics of the ear and a compression network. In 2000 Hurley et al.~\cite{Hurley2000} described an approach for ear recognition that relied on the Force Field Transform, which proved highly successful for this task. A year later, in 2001, the Forensic Ear Identification Project (FEARID) project was launched, marking the first large-scale project in the field of ear recognition~\cite{Alberink2007}.

With the beginning of the new millennium, automatic ear recognition techniques started to gain traction with the biometric community and new techniques were introduced more frequently. In 2002 Victor et al.~\cite{Victor2002} applied principal component analysis (PCA) on ear images and reported promising results. In 2005 the Scale Invariant Feature Transform (SIFT)~\cite{Hurley2005} was used for the first time with ear images, raising the bar for the performance of the existing recognition techniques. In 2006 a method based on non-negative matrix factorization (NMF) was developed by Yuan et al.~\cite{Yuan2006} and applied to occluded and non-occluded ear images with competitive results. In 2007 a method based on the 2D wavelet transform was introduced by Nosrati et al.~\cite{Nosrati2007}, followed by a technique based on log-Gabor wavelets in the same year~\cite{Kumar2007}. More recently, in 2011, local binary patterns (LBP) were used for ear-image description in~\cite{Zhi-Qin2011}, while later binarized statistical image features (BSIF) and local phase quantization (LPQ) features also proved successful for this task~\cite{Benzaoui2014, Pflug2014,Pflug2014a}. A graphical representation of the main milestones\footnote{In the opinion of the authors.} in the development of ear recognition technology (briefly discussed above) is shown in Fig.~\ref{figure:approaches_timeline}.}


\subsection{Ear Recognition Approaches}


Techniques for automatic identity recognition from ear images can in general be divided into techniques operating  on either 2D or 3D ear data. Here, we focus only on 2D approaches and refer the reader to other reviews, such as~\cite{Pflug2012}, for a detailed coverage of the field of 3D ear recognition.

A typical (2D) ear recognition system operates on images or video footage captured with commercial of-the-shelf cameras, surveillance systems, CCTV or similar everyday hardware. In a fully automatic setting, the system first detects a profile face in the input image and segments the ear from the detected region. The ear is then aligned to some predefined canonical form and processed to account for potential variability in illumination. Features are extracted from the pre-processed image and finally identity recognition is conducted based on the computed features using a suitable classification technique.
\begin{figure}[]
	\centering
	\includegraphics[width=0.87\columnwidth]{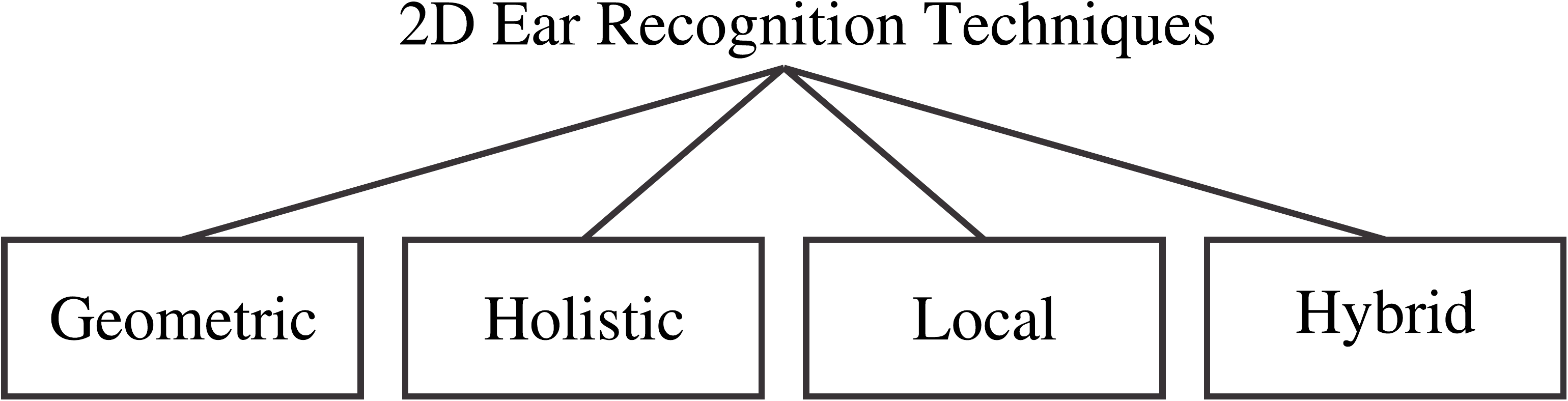}
	\caption{Taxonomy of 2D ear recognition approaches used in this paper.}
	\label{figure:tree}
\end{figure}

Depending on the type of feature extraction technique used, 2D ear recognition approaches can be divided into \textit{geometric}, \textit{holistic}, \textit{local} and \textit{hybrid} methods, as illustrated in Fig.~\ref{figure:tree}. Geometric techniques exploit the geometrical characteristics of the ear, such as the shape of the ear, locations of specific ear parts and relations between these parts. Holistic approaches treat the ear as a whole and consider representations that describe the global properties of the ear. Local approaches describe local parts or the local appearance of the ear and exploit these descriptions for recognition. The last category of hybrid approaches covers techniques that do not strictly fit into any of the other categories and usually comprises methods that combine elements from all categories or rely on multiple representations to improve performance.  

Table~\ref{Tab:_Co} presents a comparison of the techniques surveyed in this work. The techniques are ordered chronologically and labeled in accordance with our taxonomy. Information about the datasets used to evaluate the techniques as well as the reported performance is also provided. It should be noted that the results in Table~\ref{Tab:_Co} are not directly comparable with each other, as authors typically use their own experimental protocols for most datasets. The label ``Own'' in the column ``Dataset'' denotes the fact that the authors used their own dataset, while the label ``NA'' indicates that the information was not provided by the authors. The performance of the techniques is given in terms of the rank-1 recognition rate (R1), the equal error rate (EER) or the verification rate (VR)~\cite{Jain2011}.



%

\begin{table*}[!htb]
\renewcommand{\arraystretch}{1.0}
\caption{Comparative summary of 2D ear recognition techniques. The column ``Method -- Short Description'' provides a brief description of the technique used by the authors. The column ``Type'' defines the category of the approach in accordance with our taxonomy. The column ``Dataset'' presents the dataset that was used for testing, while the column ``$\#$Sub.'' stands for the number of subjects in the datasets.  The last column ``Perf.'' denotes the reported performance.}
\label{Tab:_Co}
\centering
\footnotesize
\resizebox{\textwidth}{!}{%
\begin{tabular}{ l  l lcr r}
\hline
\hline
 \textbf{Reference}                                          & \textbf{Method -- Short Description}                                                                          & \textbf{Type}               & \textbf{Dataset}                & \textbf{$\#$Sub.}  & \textbf{Perf. $[\%]$} \\ \hline\hline
1996, Burge and Burger~\cite{Burge1996}                        & Adjacency Graphs of Voronoi Diagrams                                                                  & Geometric                    &  Own                        & NA          & NA              \\ \hline
1999, Moreno et al.~\cite{Moreno1999}                          & Geometric Features | Morphological Description                                              & Geometric                    &  Own                        & 48   & 43 | 83 (R1)              \\ \hline
2000, Hurley et al.~\cite{Hurley2000}                          & Force Field Transform                                                                    & Holistic                    &  Own                        & NA         & NA              \\ \hline
2002, Victor et al.~\cite{Victor2002}                          & Principal Component Analysis -- PCA                                                                    & Holistic                    &  Own                        & 294         & 40    (R1)              \\ \hline
2003, Chang et al.~\cite{Chang2003}                            & PCA                                                                                      & Holistic                 & UND E                            & 114    &  71.6   (R1)           \\ \hline
2004, Mu et al.~\cite{Mu2004}                                  & Geometrical Measures on Edge Images                                                      & Geometric                       & USTB II                          & 77     &  85    (R1)           \\ \hline
2005, Zhang et al.~\cite{Zhang2005}                            & Independent Component Analysis -- ICA                                                     & Holistic                 & CP                & 17         & 94.1   (R1) \\
                                                                &                                                                                        &                             &  Own             & 60         & 88.3  (R1)          \\ \hline
2006, Abate et al.~\cite{Abate2006}                            & Generic Fourier Descriptor                                                               & Holistic                    &  Own                        & 70        & 96  (R1)             \\ \hline
2006, Ab.-Mottaleb et al.~\cite{Abdel-Mottaleb2006}        & Force Field and Contour Extraction                                                       & Holistic                    &  Own                        & 29         & 87.9  (R1)          \\ \hline
2006, Choras and Choras~\cite{Choras2006}                      & Geometrical Approaches on Longest Ear Contours                                          & Geometric                    &  Own                        & NA         & 100 (R1)       \\ \hline
2006, Dewi and Yahagi~\cite{Dewi2006}                          & SIFT                                                                                     & Local                       & CP                & 17         & 78.8   (R1)              \\ \hline
2006, Yuan et al.~\cite{Yuan2006}                              & Non-Negative Matrix Factorization                                                        & Holistic                       & USTB II                          & 77    & 91  (R1)               \\ \hline
2007, Arbab-Zavar et al.~\cite{Arbab-Zavar2007}                & SIFT with Pre-filtered Points                                                             & Local                       & XM2VTS                           & 63         & 91.5  (R1)             \\ \hline
2007, Kumar and Zhang~\cite{Kumar2007}                         & Log-Gabor Wavelets                                                                       & Local                    & UND                              & 113        & 90     (R1)            \\ \hline
2007, Nanni and Lumini\cite{Nanni2007}                        & Multi-Matcher                                                                           & Hybrid                 & UND E                            & 114        & 80   (R1)              \\ \hline
2007, Nosrati et al.~\cite{Nosrati2007}                        & Wavelet Transformation and PCA                                                           & Hybrid     & USTB II                          & 77         & 90.5  (R1)   \\
                                                              &                                                                                          &              & CP                               & 17         & 95.0  (R1)  \\ \hline
2007, Rahman et al.~\cite{Rahman2007}                          & Geometric Features                                                                       & Geometric                       &  Own                        & 100    & 88   (R1)              \\ \hline
2007, Yuan and Mu~\cite{Yuan2007}                              & Full-Space LDA                                            & Holistic                      & USTB II                          & 77      & 90  (R1)             \\ \hline
2008, Arbab-Zavar et al.~\cite{Arbab-Zavar2008}             & Log-Gabor Filters                                                                        & Local                    & XM2VTS                           & 63         & 85.7  (R1)             \\ \hline
2008, Choras~\cite{Choras2008}                                 & Geometry of Ear Outline                                                                  & Geometric                       &  Own                          & 188      & 86.2    (R1)           \\ \hline
2008, Dong and Mu~\cite{Dong2008}                              & Force Field Transform and NKDFA                                                        & Holistic                    & USTB IV              & 29         & 75.3    (R1)           \\ \hline
2008, Guo and Xu~\cite{Guo2008}                                & Local Binary Pattern and Cellular NN                                                             & Local                       & USTB II                 & 77        & 93.3   (R1)            \\ \hline
2008, Naseem et al.~\cite{Naseem2008}                          & Sparse Representation                                                                    & Holistic                 & UND                          & 32        & 96.9    (R1)          \\ \hline
2008, Wang et al.~\cite{Wang2008}                              & Haar Wavelet Transformation and LBP                                                      & Hybrid                       & USTB III                 & 79         & 92.4    (R1)          \\ \hline
2008, Xie an Mu~\cite{Xie2008}                                 & Enhanced Locally Linear Embedding                                                     & Holistic                 & USTB III                         & 79       & 90    (R1)             \\ \hline
2008, Zhang and Liu~\cite{Zhang2008a}                          & Null Kernel Discriminant Analysis                                                        & Holistic                 & USTB I                     & 60         & 97.7   (R1)            \\ \hline
2009, H.-Long and Z.-Chun~\cite{Hai-Long2009}                & Wavelet Transformation                                                                   & Local                       & USTB II                  & 77         & 85.7   (R1)            \\ 
                                                                &                                                                                          &                             & USTB III                & 79        & 97.2   (R1)            \\ \hline
2009, Nanni and Lumini~\cite{Nanni2009}                        & Gabor Filters                                                                            & Local                    & UND E                        & 114        & 84   (R1)              \\ \hline
2009, Xiaoyun and Weiqi~\cite{Xiaoyun2009}                     & Block Partitioning  and Gabor Transform                                          & Local              & USTB II                          & 60         & 100     (R1)           \\ \hline
2010, Alaraj et al.~\cite{Alaraj2010}                          & PCA and NN                                                                               & Holistic                  & UND              & 17          & 96   (R1)              \\ \hline
2010, Bustard and Nixon~\cite{Bustard2010}                     & SIFT                                                                                     & Local                       & XM2VTS               & 63         & 96    (R1)             \\ \hline
2010, De Marsico et al.~\cite{DeMarsico2010}                   & Partitioned Iterated Function System                                                     & Local                    & UND E                  & 114        & 59    (R1)             \\ 
                                                               &                                                                                          &                             & FERET               & 100        & 93   (R1)              \\ \hline
2010, Guiterrez et al.~\cite{Gutierrez2010}                    & Wavelet Transform and NN                                                                 & Local                   & USTB II             & 77         & 97.5    (R1)          \\ \hline
2012, Chan and Kumar~\cite{Chan2012}                           & 2D Quadrature Filter                                                                     & Local                    & IITD I                     & 125        & 96.5  (R1)            \\ 
                                                                &                                                                                          &                             & IITD II          & 221        & 96.1    (R1)          \\ \hline
2012, Kumar and Wu~\cite{Kumar2012}                            & Orthogonal Log-Gabor Filter Pairs                                                        & Local                    & IITD II                  & 221        & 95.9   (R1)          \\ \hline
2013, Baoqing et al.~\cite{Baoqing2013}                        & Sparse Representation Classification -- SRC                                                                           & Holistic                & USTB III        & 79      & \textgreater 90    (R1) \\ \hline
2013, Kumar and Chan~\cite{Kumar2013}                          & SRC of Local Gray-Level Orientations                                   & Hybrid                       & IITD I                           & 125        & 97.1   (R1)           \\ 
                                                                &                                                                                          &                             & IITD II             & 221        & 97.7   (R1)           \\ \hline
2013, Lakshmanan~\cite{Lakshmanan2013}                         & Multi-Level Fusion                                                                       & Hybrid                       & USTB II                & 77         & 99.2 (VR)         \\ \hline
2013, Prakash and Gupta~\cite{Prakash2013}                     & Enhanced SURF with NN                                                                    & Local                       & IITK I              & 190        &  2.8 (EER)       \\ \hline
2014, Basit and Shoaib~\cite{Basit2014}                        & Non-Linear Curvelet Features                                                             & Local                        & IITD II                  & 221        & 96.2   (R1)           \\ \hline
2014, Benzaoui et al.~\cite{Benzaoui2014}                      & BSIF                                                                                     & Local                      & IITD II              & 221        & 97.3   (R1)           \\ \hline
2014, Galdamez et al.~\cite{Galdamez2014}                      & Hybrid -- based on SURF, LDA and NN                                                & Hybrid                      &  Own                        & NA         & 97   (R1)              \\ \hline
2014, Jacob and Raju~\cite{Jacob2014}                          & Gray Level Co-Occurrence + LBP + Gabor Filter                              & Hybrid                      & IITD II                          & 221       & 94.1    (R1)          \\ \hline
2014, Pflug et al.~\cite{Pflug2014}                            & LPQ                                                                                      & Local                       & Several                 & 555       & 93.1    (R1)          \\ \hline
2014, Pflug et al.~\cite{Pflug2014a}                           & LPQ, BSIF, LBP and HOG -- all with LDA                                       & Hybrid             & UND J2                           & 158        & 98.7    (R1)          \\ 
                                                                &                                                                                          &                             & AMI                & 100        & 100   (R1)             \\ 
                                                                &                                                                                          &                             & IITK                  & 72         & 99.4  (R1)           \\ \hline
2014, Ying et al.\cite{Ying2014}                              & Weighted Wavelet Transform and DCT                                                       & Hybrid                      &  Own          & 75         & 98.1    (R1)          \\ \hline
2015, Benzaoui et al.\cite{Benzaoui2015}                      & LBP and Haar Wavelet Transform                                                                         & Hybrid                       & IITD               & 121        & 94.5    (R1)          \\ \hline
2015, Benzaoui et al.\cite{Benzaoui2015a}                     & BSIF                                                                                     & Local                       & IITD I                    & 125        & 96.7   (R1)           \\ 
                                                                &                                                                                          &                             & IITD II               & 221        & 97.3    (R1)          \\ \hline
2015, Bourouba et al.\cite{Bourouba2015}                      & Multi-Bags-Of-Features Histogram                                                         & Local                       & IITD I               & 125        & 96.3    (R1)           \\ \hline
2015, Meraoumia et al.\cite{Meraoumia2015}                    & Gabor Filters                                                                            & Local                    & IITD II          & 221       & 92.4     (R1)          \\ \hline
2015, Morales et al.\cite{Morales2015}                        & Global and Local Features on Ear-prints                                       & Hybrid                      & FEARID                           & 1200      & 91.3     (R1)          \\ \hline\hline
\label{table:2d}
\end{tabular}%
}
\end{table*}

\subsubsection{Geometric Approaches}

Similar to other fields of computer vision, early approaches to automatic ear recognition focused mainly on the extraction and analysis of the geometrical features of ear images. Techniques from this category are in general computationally simple and often rely on edge detection as a pre-processing step. Typically, edge information is exploited to describe the geometric properties of ears or derive geometry-related statistics that can be used for recognition. Only information pertaining to the geometry of the ear is used, which makes it easy to devise methods invariant to geometric distortions, such as rotation, scaling or even small perspective changes. On the down side, most of the texture information that could provide valuable discriminative information is discarded and not used for recognition.

\del{Pioneering work on the geometric description of ears was conducted by Burge and Burger~\cite{Burge1996}, who modeled the structure of the ears through adjacency graphs constructed from Voronoi diagrams of curve segments extracted from ear images. They also developed a graph matching procedure that could be used for authentication. An alternative to graph-based representations was introduced in~\cite{Moreno1999}, where a couple of geometric approaches were presented. The first represented each ear image with a feature vector that contained the coordinates of several characteristic ear points, while the second derived a morphology vector of the ear that captured the shape of the ear and structure of the wrinkles specific for each individual.

Many other geometric approaches were also presented in the literature and exploited: relative relationships between selected points on the ear contour~\cite{Mu2004,Rahman2007}, contour tracing techniques~\cite{Choras2006}, angle-based contour representations~\cite{Choras2006}, geometric parameterizations of the longest contour found in the ear image~\cite{Choras2006} or geometric properties of subparts of the ears~\cite{Choras2008}.}

\add{Many geometric approaches were presented in the literature and exploite: Voronoi diagrams of curve segments extracted from ear images~\cite{Burge1996}, coordinates of characteristic ear points~\cite{Moreno1999}, morphology descriptors of the ear shape and the structure of wrinkles~\cite{Moreno1999}, relative relationships between selected points on the ear contour~\cite{Mu2004,Rahman2007}, contour tracing techniques~\cite{Choras2006}, angle-based contour representations~\cite{Choras2006}, geometric parameterizations of the longest contour found in the ear image~\cite{Choras2006} or geometric properties of subparts of ears~\cite{Choras2008}.}

One limitation of geometric approaches is their dependency on edge detectors, which are known to be susceptible to illumination variations and noise. Many techniques also require exact locations of specific ear points that may be difficult to detect in poor quality \del{or low-resolution} images \add{or when an ear is occluded~\cite{Pflug2012,Yan2007}}.

\subsubsection{Holistic Approaches}

Holistic approaches differ conceptually from geometric techniques discussed in the previous section. Instead of geometric properties, holistic approaches exploit the global appearance of the ear and compute representations from input images that encode the ear structure as a whole. As the appearance of an ear changes significantly with pose or illumination changes, care needs to be taken before computing holistic features from images and normalization techniques need to be applied to correct for these changes prior to feature extraction. Holistic techniques were reported to provide competitive performance on well aligned and well preprocessed images, but may experience difficulties if this is not the case.

One popular (holistic) approach to ear recognition was presented in~\cite{Hurley2000}, where the so-called Force Field Transform was introduced. The technique computes a force field from the input ear image by treating the pixels as sources of a Gaussian force field. The force field is then analyzed and its properties are exploited to compute similarities between ear images. Force-field-transform-based techniques proved very successful and were extended several times, e.g., in~\cite{Hurley2005,Abdel-Mottaleb2006,Dong2008,Hurley2002}.

Linear and nonlinear subspace projection techniques have also been introduced to the field of ear recognition. With subspace projection techniques, ear images are represented in the form of weights of a linear combination of some basis vectors in either the input (pixel) space or some high-dimensional Hilbert space. Existing examples of techniques from this group include: principal component analysis (PCA)~\cite{Victor2002,Chang2003,Alaraj2010}, independent component analysis (ICA)~\cite{Zhang2005}, non-negative matrix factorization (NMF)~\cite{Yuan2006}, full space linear discriminant analysis (LDA)~\cite{Yuan2007}, enhanced locally linear embedding (ELLE)~\cite{Xie2008}, or null kernel discriminant analysis (NKDA)~\cite{Zhang2008a}. A similar idea is also exploited with sparse representations of ear images, where images are represented in the form of a sparse linear combination of the training images (or some other dictionary)~\cite{Naseem2008,Baoqing2013}.

Holistic techniques operating in the frequency domain have been used for ear recognition as well. One such example is the technique of Abate et al.~\cite{Abate2006}, which computes a global image descriptor in the frequency domain to represent ear images and is able to achieve rotation invariance by representing the ear image in polar coordinates prior to descriptor computation.

\subsubsection{Local Approaches}\label{subsec: local}

Local approaches for ear recognition extract features from local areas of an image and use these features for recognition. Different from geometric approaches, local approaches do not rely on the locations of specific points or relations between them, but on the description of the local neighborhood (or area) of some points in the image. The points of interest to local approaches must not necessary correspond to structurally meaningful parts of the ear, but can in general represent any point in the image. Two types of techniques can be considered local in this context: \textit{i)} techniques that first detect keypoint locations in the image and then compute separate descriptors for each of the detected keypoints~\cite{neurocomputing6}, and \textit{ii)} techniques that compute local descriptors densely over the entire image.

Local techniques (together with hybrid approaches) currently achieve the most competitive results in the field of ear recognition. Unlike geometric or holistic approaches, local techniques mostly encode texture information~\cite{neurocomputing6}. Many of the techniques from this category exhibit some level of robustness to small image rotations and alignment errors owing to the fact that histograms are computed over smaller image areas for most descriptors. The computational complexity of many low-level descriptors, such as local binary patterns (LBPs), is low and the descriptors can be computed extremely efficiently. However, others, such as Gabor features, for example, are computationally more demanding.

Examples of techniques that first detect keypoint locations in ear images and then extract SIFT (scale invariant feature transform) descriptors around these keypoints were presented in~\cite{Dewi2006,Arbab-Zavar2008} and~\cite{Bustard2010}. An approach similar in nature albeit based on speeded up robust features (SURF) was presented in~\cite{Prakash2013}. A desirable characteristic of this group of techniques is the independent description of keypoint locations, which makes partial matching of the extracted descriptors possible and allows designing methods that exhibit robustness to partial occlusions of the ear. On the other hand, information about the global ear structure that could help with the recognition process is typically lost with this group of methods.

Techniques from the second group compute local descriptors in a dense manner without first detecting keypoints in an image. The resulting image representation is comprised of local descriptors that still encode the global structure of the image, but at the expense of loosing robustness to partial occlusions (in most cases). Typical examples of techniques from this group rely on: Gabor or log-Gabor filters and other types of wavelet representations~\cite{Kumar2007,Kumar2012,Arbab-Zavar2008,Hai-Long2009,Nanni2009,Xiaoyun2009,Gutierrez2010,Chan2012,Meraoumia2015}, curvelet representations~\cite{Basit2014}, dense SIFT (DSIFT)~\cite{Krizaj2010}, local binary patterns (LBP)~\cite{Guo2008}, local phase quantization features (LPQ)~\cite{Pflug2014} and rotation invariant LPQs (RILPQ)~\cite{OjansivuRILPQ2008}, histograms of oriented gradients (HOG)~\cite{Pflug2014a,DamerHOG2012}, patterns of oriented edge magnitudes {POEM)~\cite{Vu2010} or learned local descriptors such as the binarized statistical image features (BSIF)~\cite{Benzaoui2014,Benzaoui2015a}. Other methods that also fall into this group are based on fractals~\cite{DeMarsico2010} or the bag-of-features model~\cite{Bourouba2015}.

\subsubsection{Hybrid Approaches}

The last category of methods in our taxonomy covers methods that combine elements from other categories or use multiple representations to increase the recognition performance. 

One such approach was introduced by Nosrati et al.~\cite{Nosrati2007} and combined wavelets and PCA. Another technique, presented in~\cite{Wang2008}, used LBPs on images transformed with the Haar wavelet transform. Kumar and Chan~\cite{Kumar2013} adopted the sparse representation classification algorithm and applied it to local gray-level orientation features, while Galdamez et al.~\cite{Galdamez2014} first computed SURF features from the images and then reduced the dimensionality using LDA. A combination of the wavelet transform and the discrete cosine transform (DCT) was presented in~\cite{Ying2014} and a hybrid method based on LBPs and the Haar transform was introduced in~\cite{Benzaoui2015}. Various combinations of local descriptors and subspace projection techniques were assessed in~\cite{Pflug2014a}.

A hybrid technique trying to improve performance by using multiple matchers was presented by Nanni and Lumini in~\cite{Nanni2007}. Lakshmanan described an approach based on multi-level fusion in~\cite{Lakshmanan2013}. Jacob and Raju~\cite{Jacob2014} combined gray-level co-occurrence Matrices, LBPs and Gabor filters for ear recognition, while Morales et al.~\cite{Morales2015} tried to exploit global and local features to reliably recognize ear-prints.

Hybrid ear recognition techniques are often quite competitive, but depending on their building blocks may also be computationally more complex than simpler holistic or local techniques.
\del{\subsection{Characteristics of Ear Recognition}}

\mvf{Ear images needed for the recognition process can typically be extracted from profile head shots or video footage. The acquisition procedure is \textit{contactless} and \textit{nonintrusive} and also \textit{does not depend on the cooperativeness} of the person one is trying to recognize. In this regard ear recognition technology shares similarities with other image-based biometric modalities, such as facial images, periocular biometrics or gait patterns. These characteristics  make it a suitable modality for security and surveillance applications, where recognition from a distance is a must.

Another appealing property of ear biometrics is its \textit{distinctiveness}. Recent studies empirically validated Iannarelli's observations that certain features of the ear are distinct even for identical twins~\cite{Sim2012ICPR}. This fact has significant implications for security related applications and puts ear images on par with epigenetic biometric modalities, such as the iris.

Last but not least, ear images can serve as \textit{supplements} for other biometric modalities in automatic recognition systems. In surveillance applications, for example, where face recognition technology may struggle with profile faces, the ear can provide important cues on the identity of people in the surveillance footage. The importance and potential value of ear recognition technology for multi-modal biometric systems is also evidenced by the number of research studies on this topic, e.g.,~\cite{MM2007,MM2008,MM2008b,MM2009,MM2009b}.}

\del{ 
\section{State-of-the-Art in Descriptor-Based Ear Recognition}
\label{section:state-of-the-art}

In this section we present an in-depth description of the most promising descriptor-based techniques, which offer a reasonable trade-off between computational complexity and recognition performance. All techniques reviewed in this section are also used in our experiments in Section~\ref{section:experiments}.

\subsection{Local Binary Patterns}

Local Binary Patterns (LBP) represent powerful texture descriptors that achieved competitive recognition performance in various areas of computer vision~\cite{Pietikainen2011}, as an ear recognition technique and as a combination with other techniques, e.g.,~\cite{Pflug2014a,Guo2008,Benzaoui2015}. LBPs encode the local texture of an image by generating binary strings from circular neighborhoods of points thresholded at the gray-level value of their center pixels. The generated binary strings are interpreted as decimal numbers and assigned to the center pixels of the neighborhoods. In practice only binary strings with at most two bitwise transitions from 0 to 1 (or vice versa) are considered in the final descriptor. Most methods exploiting LBPs with a 8-pixel neighborhood for texture description, compute 59-bin histograms from local image blocks and then concatenate the computed histograms over all blocks into a global texture descriptor that can be used for recognition. A similar procedure is also used in our experiments in Section~\ref{section:experiments}.

\textit{Advantages:} computational simplicity, the fact that the texture of the ear is highly discriminative.
\subsection{(Rotation Invariant) Local Phase Quantization}

Local Phase Quantization (LPQ) features~\cite{Ojansivu2008} are very similar in essence to LBPs, as the local image texture is encoded using binary strings, histograms are computed from the binary strings of local image blocks and concatenated into the final representation of the given image. LPQ features are computed from the Fourier phase spectrum of an image: the local neighborhoods of every pixel in the image are first transformed into the frequency domain using a short-term Fourier transform. Local Fourier coefficients are computed at four selected frequency points and the local phase information contained in these (complex) coefficients is then encoded. Here, a similar quantization scheme is used as in iris recognition systems, where every complex Fourier coefficient contributes two bits to the final binary string. The result of this coding procedure is a 8-bit binary string for every pixel in the image from which the local 256-bin histograms are computed and later concatenated into a global descriptor of the image.

An extension of this technique to rotation invariant local phase quantization features (RILPQ) was presented in~\cite{OjansivuRILPQ2008}. The idea here is very similar to the original LPQ technique with the difference that a characteristic orientation is first estimated for the given local neighborhood and then this orientation is used to compute a directed version of the binary descriptor. The binary descriptor is computed with the same procedure as the original LPQ, but  every local neighborhood is first rotated in accordance with its characteristic orientation.

\textit{Advantages:} blur invariant (LPQ, RILPQ) and rotation invariant (RILPQ).
\subsection{Binarized Statistical Images Features}
In Binarized Statistical Images Features (BSIF)~\cite{Kannala2012} binary strings (encoding texture information) are again constructed for each pixel in the image, but this time by projecting image patches onto a subspace, whose basis vectors are learned from natural images. The subspace coefficients are then binarized using simple thresholding. This procedure is equivalent to filtering the input image with a number of pre-learned filters and binarizing the filter responses at each pixel location. Each filter contributes 1 bit to the binary string of a pixel making the length of the binary string dependent on the number of filter used. Similar to LBP and LPQ, the binary string of each pixel is interpreted in decimal form and a global histogram-based representation is constructed for the given images by concatenating histograms constructed from smaller image blocks. The use of BSIF features for ear recognition was advocated by Pflug et al. in~\cite{Pflug2014a}, where excellent performance was reported. In the experimental section, we implement a basic BSIF-based recognition technique and again observe competitive performance.

\textit{Advantages:} binary strings are not constructed based on heuristic operations, but on the basis of statistics of natural images.
\subsection{Histograms of Oriented Gradients}

%
Descriptors exploiting Histograms of Oriented Gradients (HOG) were originally introduced for the problem of human detection by Dalal and Triggs~\cite{DalaHOG2005}, but have since  been successfully applied to various fields of computer vision, including ear recognition~\cite{Pflug2014a,DamerHOG2012}. The computation starts by calculating the gradient of the image using 1-dimensional convolutional masks, i.e., $[-1, 0, 1]$ and $[-1, 0, 1]^T$. The image is then divided into a number of cells and compact histograms of quantized gradient orientations are computed for each cell. Here, a voting procedure is used during histogram construction, so that pixels with higher gradient magnitudes contribute more to the histogram bins than pixels with lower magnitudes. Neighboring cells are then grouped into larger blocks and normalized to account for potential changes in contrast and illumination. This normalization procedure is applied in a sliding-window manner over the entire image with some overlap between neighboring blocks. Ultimately, all normalized histograms are concatenated into the final HOG descriptor that can be used for matching and recognition.

\textit{Advantages:} robust towards moderate illumination changes.

\subsection{Dense Scale Invariant Feature Transform}

The original approach to Scale Invariant Feature Transform (SIFT) calculation, introduced by Lowe in~\cite{Lowe2004}, includes both a keypoint detector, capable of finding points of interest in an image, as well as a local descriptor. Early techniques to ear recognition relied on the SIFT keypoint detector as well as the SIFT descriptor, e.g.,~\cite{Dewi2006,Arbab-Zavar2008,Bustard2010}, and, therefore, demonstrated a high degree of robustness towards partial occlusions. However, more recent techniques, compute dense SIFT (DSIFT) representations from the images and do not rely on the keypoint detector, but arrange keypoints uniformly into a grid over the image (e.g.,~\cite{MoralesSIFT2013,Krizaj2010}). The SIFT descriptor shares similarities with the HOG descriptor. For every point of interest, SIFT considers a local neighborhood of $16\times16$ pixels. This neighborhood is partitioned into sub-regions of $4\times4$ pixels and for each sub-region an 8-bin histogram is computed based on the orientations and magnitudes of the image gradient in that sub-region. The gradients are also weighted by a Gaussian function to give more importance to image gradients closer to the point of interest and normalized by the dominant gradient orientation to achieve rotation invariance. The final dimensionality of the SIFT descriptor is 128 for a single keypoint, so care needs to be taken when computing DSIFT representations from the image. The dimensionality of final feature vector can easily become computationally prohibitive if too many grid points are chosen for DSIFT calculation.

\subsection{Gabor Wavelets}

2D Gabor wavelets were originally introduced by Daugman~\cite{Daugman1985} for the problem of iris coding, but due to their ability to analyze images at multiple scales and orientations, they have been successfully employed in other problem areas as well~\cite{StrucGabor2009,Struc2009,Struc2010}. To extract Gabor features from an image, the image is convolved with the entire family of Gabor wavelets (filters), the magnitude responses of the convolution outputs are retained (the phase responses are discarded), down-sampled and concatenated into a global feature vector encoding multi-resolution, orientation-dependent texture information of the input image. Techniques based on the outlined procedure and its modifications (e.g., using log-Gabor wavelets) are among the most popular techniques for ear recognition~\cite{Kumar2007,Kumar2012,Nanni2009,Xiaoyun2009,Meraoumia2015}.
\textit{Advantages:} excellent discriminative properties.
\textit{Disadvantages:} computational complexity.

\subsection{Patterns of Oriented Edge Magnitudes}
%
%

Patterns of Oriented Edge Magnitudes (POEM)~\cite{Vu2010} represent another popular approach to texture description that combines ideas from LBP and HOG descriptors as well as Gabor wavelets. The POEM construction procedure starts by computing the gradient of the input image and building magnitude-weighted histograms of gradient orientations for every pixel in the image. This histogram is computed from local pixel neighborhoods referred to by the authors as cells. As opposed to HOG, POEM computes the histograms densely in a sliding window-manner over the entire image. After this step, every pixel in the image is represented by a local histogram of quantized gradient orientations, or in other words, the image is decomposed into $m$ oriented gradient images, where $m$ is the number of discrete orientations of the local histograms. Each of these images is then encoded using the LBP operator and a global image descriptor is constructed by concatenating all block histograms computed from the oriented gradient images.

\textit{Advantages:} orientational-selectivity, robustness to moderate illumination changes and low-computational complexity.

} 

\section{Existing Datasets}
\label{section:data}

\begin{figure}[!t]
	\centering
	\includegraphics[width=\columnwidth]{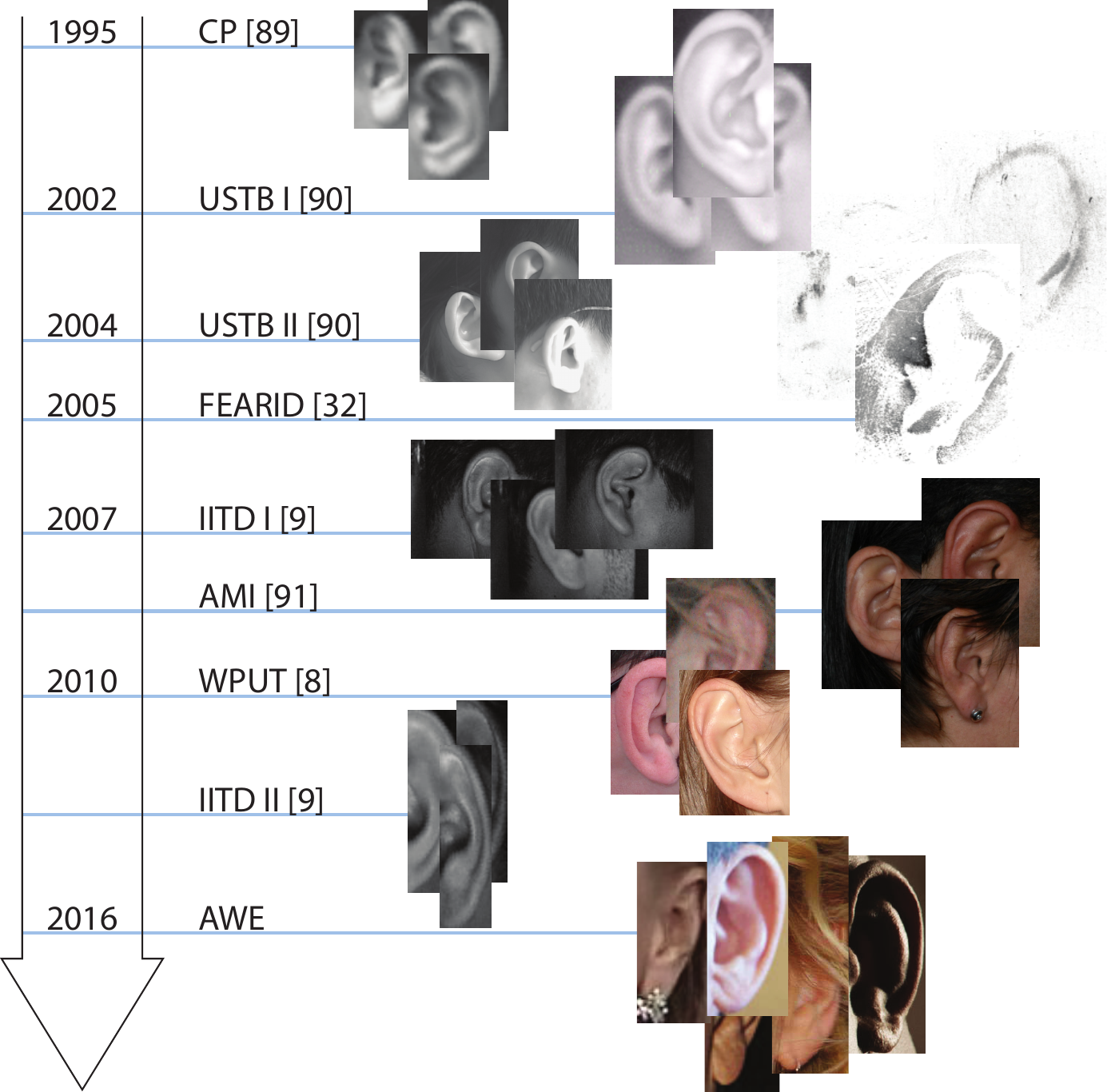}
	\caption{Sample images of available ear datasets. The figure also shows the development of the datasets through time. Note how the datasets have progressed towards more realistic imaging conditions.}
	\label{figure:db_timeline}
\end{figure}
\del{In this section we present an overview of existing ear datasets that can be used for training and evaluation of 2D ear recognition approaches. We specifically focus on datasets of cropped ear images suitable for studying ear recognition rather than ear detection approaches. We also discuss profile-face datasets that can be used for both ear detection and recognition, but review these more briefly at the end of the section. }In this section we present an overview of the existing ear datasets that can be used for training and evaluation of 2D ear recognition approaches. We specifically focus on datasets of cropped ear images suitable for studying ear recognition rather than ear detection approaches. A comparative summary of the most popular ear datasets is provided in Table~\ref{table:ears1}. The datasets have different characteristics and the corresponding ear images exhibit different levels of variability, as also illustrated in Fig.~\ref{figure:db_timeline}. 

\begin{table*}[]
\centering
\caption{Comparative summary of the existing ear datasets and their characteristics. The category ``Sides'' evaluates whether images from the left or right side of the head are present in the dataset, ``Yaw'' and ``Pitch'' provide information about the ear rotation, ``Occlusions'' and ``Accessories'' assess whether occlusion and accessories are visible in the images and the last two categories ``Gender'' and ``Ethnicities'' indicate whether both sexes are present and what kind of racial variation is accounted for in the datasets.}
\label{table:ears1}
\small
\resizebox{\textwidth}{!}{%
\begin{tabular}{l|llllllllll}
DB                & Year & \#Subjects & \#Images & Sides & Pitch  & Yaw    & Occlusions & Accessories & Gender & Ethnicities \\ \hline
CP~\cite{CP_dataset1995} & 1995 & 17         & 102      & left  & none   & none   & none       & none        & both    & white       \\
USTB I~\cite{USTBdatasets}            & 2002 & 60         & 185      & right & mild   & none   & none       & yes         & both    & asian       \\
USTB II~\cite{USTBdatasets}          & 2004 & 77         & 308      & right & mild   & mild   & none       & yes         & both    & asian       \\
IITD I~\cite{Kumar2012}           & 2007 & 125        & 493      & right & severe   & none   & none       & yes         & both    & asian       \\
AMI~\cite{AMI_dataset2008}              & NA & 100        & 700      & both  & severe & mild   & mild       & none        & both    & white       \\
WPUT~\cite{Frejlichowski2010}             & 2010 & 501        &  2071     & both  & severe & mild   & severe     & yes         & both    & white       \\
IITD II~\cite{Kumar2012}         & NA & 221        & 793      & right & none   & none   & none       & none        & both    & asian       \\
\dname               & 2016 & 100        & 1000     & both  & severe & severe & severe     & yes         & both    & various     \\
\end{tabular}%
}
\end{table*}

\subsection{CP Ear Dataset}

The Carreira-Perpinan (CP) ear dataset~\cite{CP_dataset1995} represents one of the earliest publicly available datasets for ear recognition. Introduced in 1995, the dataset contains 102 ears of  17 distinct subjects. All images have been captured in controlled conditions, so most of the image variability is due to minor pose variations and, of course, subject identity. A few sample images from the dataset are also shown at the top of Fig.~\ref{figure:db_timeline}. 


\subsection{ IITD Ear Dataset}
The ear dataset of the Indian Institute of Technology Delhi consists of two sub-datasets~\cite{Kumar2012}. The first contains $493$ gray-scale images of $125$ subjects and the second $793$ images of $221$ subjects. All images were taken at different indoor lightning conditions and from approximately the same profile angle. The first dataset (IITD I) is available in raw and pre-processed form, while the second (IITD II) is available only in the pre-processed form. With pre-processing the authors of the datasets ensured that: \textit{i)} all ears are tightly cropped, \textit{ii)} all images are of equal dimensions, and \textit{iii)} all ears are centered and mutually aligned. Additionally, all images of the left ear are mirrored, so the entire dataset appears to consist of images of the right ear. The number of images per subject ranges from 3 to 6. No major occlusions are present. The dataset is available per request. 

\subsection{ USTB Ear Datasets}	
The University of Science \& Technology Beijing introduced four distinct ear datasets enjoying a high level of popularity within the community~\cite{USTBdatasets}. The first two datasets, USTB I and USTB II, contain only cropped ear images, while datasets III and IV feature complete head profile shots. In the first dataset the authors captured $3$ to $4$ images of $60$ volunteers resulting in a total of $185$ images. For the second dataset the authors acquired $4$ images of $77$ volunteers, altogether $308$ images. In both datasets indoor lightning was used, but the second dataset contains more loosely cropped images, a higher level of illumination-induced variability and pitch angles between $\pm30\degree$ making it more challenging. The USTB III and USTB IV datasets 
contain full face-profile images captured under specific angles and occlusions. USTB III contains $1600$ images of $79$ subjects and USTB IV contains images of $500$ subjects with everyone photographed at $17$ different angles with $15\degree$ steps in between. The USTB datasets are available per request and are free for research purposes. 

\subsection{ AMI Ear Dataset}
The AMI (translated from Spanish as Mathematical Analysis of Images) ear dataset~\cite{AMI_dataset2008} was collected at the University of Las Palmas and consists of 700 images of a total of 100 distinct subjects in the age group of $19$--$65$ years. All images were taken under the same illumination and with a fixed camera position. For each subject 1 image of the left ear and 6 images of the right ear were taken. All ears in the dataset are loosely cropped, so that the ear area covers approximately one half of the image. The poses of subjects vary slightly in yaw (all images are still from profile nonetheless) and severely in pitch (subject looking up at $45\degree$ angle) angles. The dataset is publicly available. 

\subsection{ WPUT Ear Dataset}
The ear dataset of the West Pommeranian University of Technology (WPUT)~\cite{Frejlichowski2010} was introduced in 2010 with $2071$ ear images belonging to $501$ subjects. However, the dataset currently available for download contains $3348$ images of $471$ subjects, with $1388$ duplicates. Images belonging to subject IDs from $337$ to $363$ are missing. The dataset contains annotation according to the following categories: gender, age group, skin color, head side, two types of rotation angles, lightning conditions, background (cluttered or heterogeneous), and occlusion type (earrings, hat, tattoos etc.). There are between $4$ and $10$ images per subject in the dataset. Images were taken under different indoor lightning conditions and (head-)rotation angles ranging from approximately $90\degree$ (profile) to $75\degree$. 
The dataset is publicly available free of charge. 

\del{\subsection{ FEARID Dataset}}
\mvf{The FEARID dataset~\cite{Alberink2007} was collected as part of the FEARID project and differs from other datasets in they type of data it contains. Unlike other datasets, FEARID does not contain ear images, but ear-prints, which were acquired from different research groups using specific scanning hardware. Ear-print data differs significantly from regular images, as there are no occlusions, no variable angles and no illumination variations. However, other sources of variability influence the appearance of the ear-prints, such as the force with which the ear was pressed against the scanner, scanning-surface cleanliness and other similar factors. The acquisition protocol for the dataset was designed to simulate the appearance of ear-prints that would normally be found at crime scenes and resulted in a dataset comprising $7364$ images of $1229$ subjects. The FEARID dataset is interesting, because it was collected with forensic applications in mind  and with a somehow different purpose than other ear datasets popular within the biometric community.}  

\subsection{\del{Face Profile}\add{Other} Datasets}
In addition to the datasets reviewed above, datasets of ear-prints and profile-face images are also suitable for studying ear recognition technology. While there are many such datasets publicly available, we discuss (in this section) only datasets that were used for ear recognition in the literature.

\mvt{The FEARID dataset~\cite{Alberink2007} was collected as part of the FEARID project and differs from other datasets in the type of data it contains. Unlike other datasets, FEARID does not contain ear images, but ear-prints, which were acquired from different research groups using specific scanning hardware. Ear-print data differs significantly from regular images, as there are no occlusions, no variable angles and no illumination variations. However, other sources of variability influence the appearance of the ear-prints, such as the force with which the ear was pressed against the scanner, scanning-surface cleanliness and other similar factors. The acquisition protocol for the dataset was designed to simulate the appearance of ear-prints that would normally be found at crime scenes and resulted in a dataset comprising $7364$ images of $1229$ subjects. The FEARID dataset is interesting because it was collected with forensic applications in mind  and with a somehow different purpose than other ear datasets popular within the biometric community.}

The first dataset from \del{this group}\add{the group of profile-face-image datasets} is the dataset of the Indian Institute of Technology Kanpur (IITK)~\cite{Prakash2012}. During data collection, several face images with multiple fixed head poses were captured and organized into three subsets: \textit{i)} the first subset contains 801 profile-only images of $190$ subjects, \textit{ii)} the second subset contains $801$ images of $89$ subjects with pitch angle variations, and \textit{iii)} the third subset contains $1070$ images of the same 89 subjects, but with yaw angle variations.

The UBEAR dataset~\cite{ubear}  
comprises $4429$ profile images from $126$ subjects taken from both the left and the right side. The images  were captured under different illumination conditions. Partial occlusions of the ear are also present on some of the images. This dataset is interesting because images were captured while the subjects were moving. This characteristic is useful for studying techniques for video-based ear recognition where blurring and shearing effects typically appear.

The datasets of the University of Notre Dame (UND)\footnote{\url{https://sites.google.com/a/nd.edu/public-cvrl/data-sets}}
contain multiple subsets with 2D and 3D ear data. Subset E contains  2D images of the left side of the face of $114$ subjects. Subsets F, G, J2 and NDOff-2007 all contain images with clearly visible ears and corresponding 3D and 2D data with $942$, $738$, $1800$ and $7398$ images, respectively. These subsets are useful not only for 2D or 3D ear detection and recognition, but also enable direct comparison of 2D and 3D approaches.

The FERET~\cite{FERET2000} dataset in its final version contains $14,051$ images. The dataset is interesting because images contain annotations about the subjects' facial expression (neutral or not) and  pose angles, which can be interesting covariates to study in ear recognition.

The Pose, Illumination and Expression (PIE)~\cite{cmupie} dataset from the Carnegie Mellon University contains images of $68$ subjects. The dataset comprises images captured in $13$ different poses, $43$ illumination conditions and $4$ different facial expressions for each of the 68 subjects resulting in a total of $40.000$ images for the entire dataset. 

The XM2VTS~\cite{Messer1999} 
dataset features frontal and profile face images as well as video recordings of $295$ subjects. The dataset was captured in controlled conditions during eight recording sessions and over a period of around five months. 

Another dataset with video recordings is the ear dataset of the West Virginia University (WVU). This dataset is interesting because the authors used advanced capturing procedures (described in~\cite{Fahmy2006}) that allowed them to precisely record ears from different angles. The dataset contains images/videos of $137$ subjects. 


\section{\dname Dataset and Toolbox}\label{Section:DatasetAndToolbox}

Most datasets reviewed in the previous section were captured in controlled, laboratory conditions.
In this section we introduce the Annotated Web Ears (\dname) dataset, which differs significantly in this regard from other existing datasets. We also present the \dname Matlab toolbox for ear recognition. The dataset and toolbox are both publicly available from \url{http://awe.fri.uni-lj.si}.

\subsection{The \dname Dataset}

\textbf{Motivation:}
Most existing datasets for evaluating ear recognition technology contain images captured in controlled laboratory settings or profile shots of people posing outdoors in front of a camera. The variability of these images is usually limited and the sources of variability are pre-selected by the authors of the datasets. This is in stark contrast to the field of face recognition, where in the last decade the research focus shifted to datasets of completely uncontrolled  images, which helped to advance the state-of-technology significantly. A cornerstone in this development were datasets gathered directly from the web, such as the LFW~\cite{LFWTech}, PubFig~\cite{KumarPuFig2009}, FaceScrub~\cite{Ng2014}, Casia WebFaces~\cite{CasiaWebFaces2014}, IJB-A~\cite{KlareIJBA2009} and others. These datasets introduced the notion of \textit{images captured in the wild}, with the wordplay indicating that the images were acquired in  uncontrolled environments. The difficulty of these datasets provided significant room for improvements in the technology and spurred the development of the field over recent years.

By today, the reported performance of ear-recognition techniques has surpassed the rank-1 recognition rate of 90\% on most available datasets. This fact suggests that the technology has reached a level of maturity that
easily handles images captured in laboratory-like settings and more challenging datasets are needed to identify open problems and provide room for further advancements. Inspired by the success of datasets collected \textit{in the wild} for face recognition, we gather a new dataset of ear images from the web and make it available to the research community. 

\textbf{Dataset collection:} All images of the \dname dataset were collected from the web by first compiling a list of people, for whom it was reasonably to assume that a large amount of images could be found online. Similar to other datasets collected in the wild this list mostly featured actors, musicians, politicians and the like. A web crawler was then used to gather large amounts of images for each person from the list by sending appropriate search queries to Google's image search\footnote{In most cases the name and surname generated enough hits.}. To avoid pre-filtering of the images with automatic detection techniques~\cite{KlareIJBA2009} and, thus, limit the image variability, all images were manually inspected and 10 images per subject were selected for inclusion in the dataset.

\textbf{Structure and annotations:} In total, the dataset contains $1000$ ear images of $100$ subjects. Each subject has $10$ images of a different quality and size. The smallest image in the dataset has $15\times 29$ pixels, the biggest $473\times 1022$ pixels and the average image size is $83\times 160$ pixels. The ear images are tightly cropped and do not contain profile faces as shown in Fig.~\ref{figure:awed_sample}.

\begin{figure}[]
\captionsetup{type=figure}
	\center
		\subfloat{
			\includegraphics[height=0.2\columnwidth]{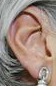}
		}
		\subfloat{
			\includegraphics[height=0.2\columnwidth]{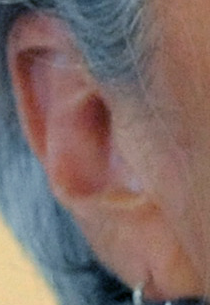}
		}
		\subfloat{
			\includegraphics[height=0.2\columnwidth]{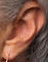}
		}
		\subfloat{
			\includegraphics[height=0.2\columnwidth]{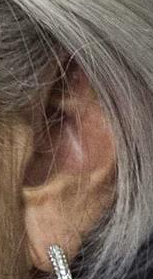}
		}
		\subfloat{
			\includegraphics[height=0.2\columnwidth]{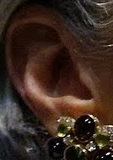}
		}
		\subfloat{
			\includegraphics[height=0.2\columnwidth]{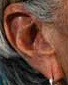}
		}\\
		\subfloat{
			\includegraphics[height=0.2\columnwidth]{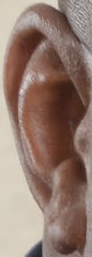}
		}
		\subfloat{
			\includegraphics[height=0.2\columnwidth]{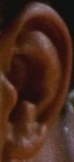}
		}
		\subfloat{
			\includegraphics[height=0.2\columnwidth]{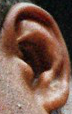}
		}
		\subfloat{
			\includegraphics[height=0.2\columnwidth]{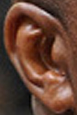}
		}
		\subfloat{
			\includegraphics[height=0.2\columnwidth]{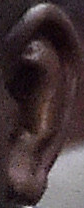}
		}
		\subfloat{
			\includegraphics[height=0.2\columnwidth]{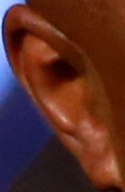}
		}
		\subfloat{
			\includegraphics[height=0.2\columnwidth]{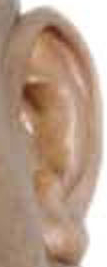}
		}
		\subfloat{
			\includegraphics[height=0.2\columnwidth]{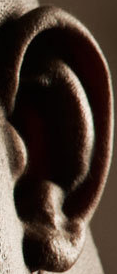}
		}
		\caption{Sample images from the Annotated Web Ears (AWE) dataset. Images in each row correspond to one subject from the dataset. Note the extend of variability present in the images.}
		\label{figure:awed_sample}
\end{figure}

Each image in the dataset was annotated by a trained researcher according to the following categories: gender, ethnicity, accessories, occlusion, head pitch, head roll, head yaw, and head side. The pitch, roll and yaw angles were estimated from the head pose. The available labels for each category are shown in Table~\ref{Tab:_labels} and the distribution of the labels for the entire dataset in Fig.~\ref{img:db_1000}. Additionally, the location of the tragus was also marked in each image.

\begin{table}[tb]
\begin{threeparttable}
\renewcommand{\arraystretch}{1.05}
\caption{Annotation categories and corresponding labels for Annotated Web Ears (\dname) dataset.}
\label{Tab:_labels}
\small
\centering
\begin{tabular}{ l|  c}
Category    &   Available labels\\
\hline
\hline
Gender      & Male, Female \\
\hline
\multirow{ 2}{*}{\small Ethnicity}    & White, Asian, South Asian, Black\\
                                      & Middle Eastern, South American, Other \\
\hline
Accessory & None, Earrings, Other \\
\hline
Occlusion*   & None, Mild, Severe \\
\hline
Head Pitch**       & Up ++, Up +, Neutral, Down +, Down ++ \\
\hline
\multirow{ 2}{*}{\small Head Roll**}        & To Right ++, To Right +, Neutral\\ & To Left +, To Left ++ \\
\hline
\multirow{ 2}{*}{\small Head Yaw}          & Frontal Left, Middle Left, Profile Left\\
                                            & Profile Right, Middle Right, Frontal Right \\
\hline
Head Side   & Left, Right \\
\end{tabular}
\begin{tablenotes}
\footnotesize
\item \add{* \textit{mild} denotes occlusions, where small parts of the ear are covered by earrings or hair strands, \textit{severe} denotes occlusions where larger parts of the ear are not visible and covered by hair or other objects\\ ** $++$ means larger than $10\degree$, $+$ means between $2\degree$ and $10\degree$}
\end{tablenotes}
\end{threeparttable}
\end{table}

\begin{figure*}[!htb]
\captionsetup{type=figure}
	\center
		\subfloat[Gender]{\makebox[0.23\textwidth][c]{
			\includegraphics[width=0.23\textwidth]{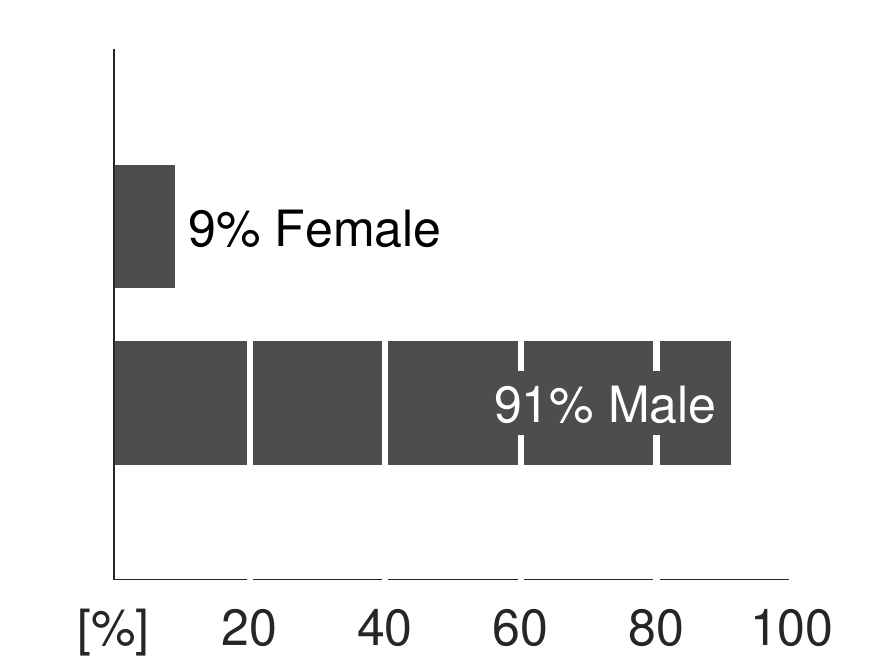}
			\label{img:db_sex}
		}}\hfill
		\subfloat[Ethnicity]{\makebox[0.23\textwidth][c]{
			\includegraphics[width=0.23\textwidth]{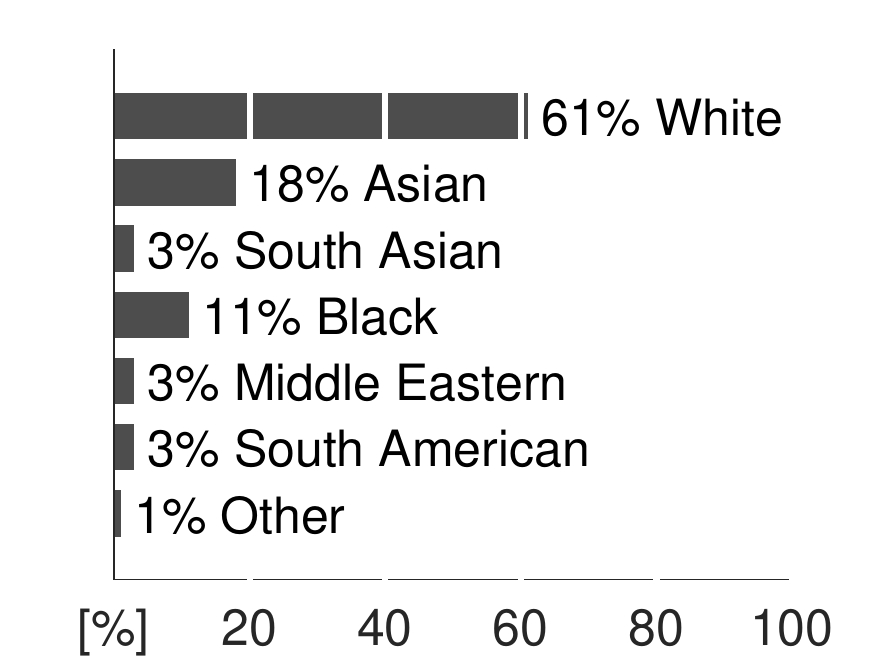}
			\label{img:db_ethnicity}
		}}\hfill
		\subfloat[Accessories]{\makebox[0.23\textwidth][c]{
			\includegraphics[width=0.23\textwidth]{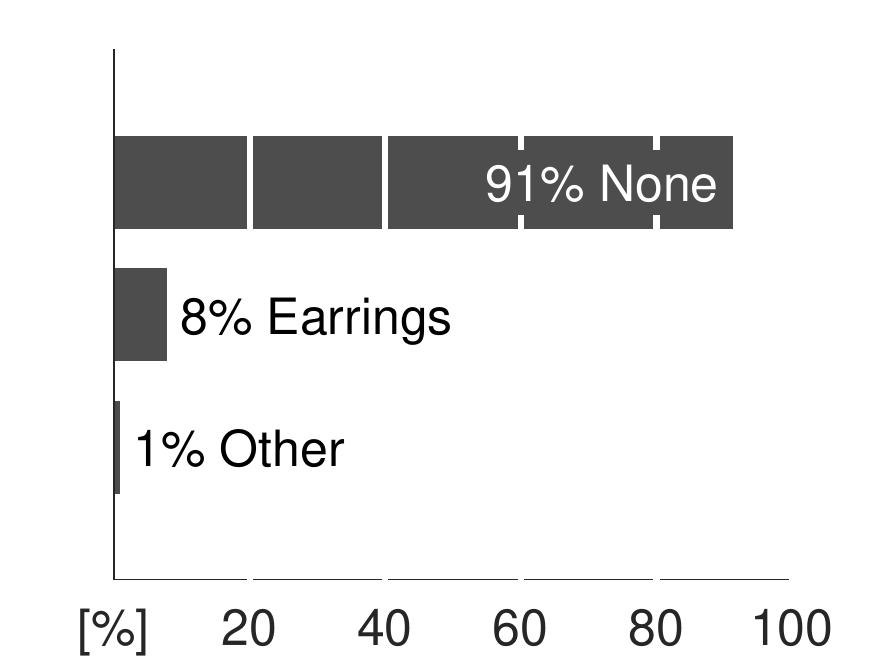}
			\label{img:accessories}
		}}\hfill
		\subfloat[Occlusion]{\makebox[0.23\textwidth][c]{
			\includegraphics[width=0.23\textwidth]{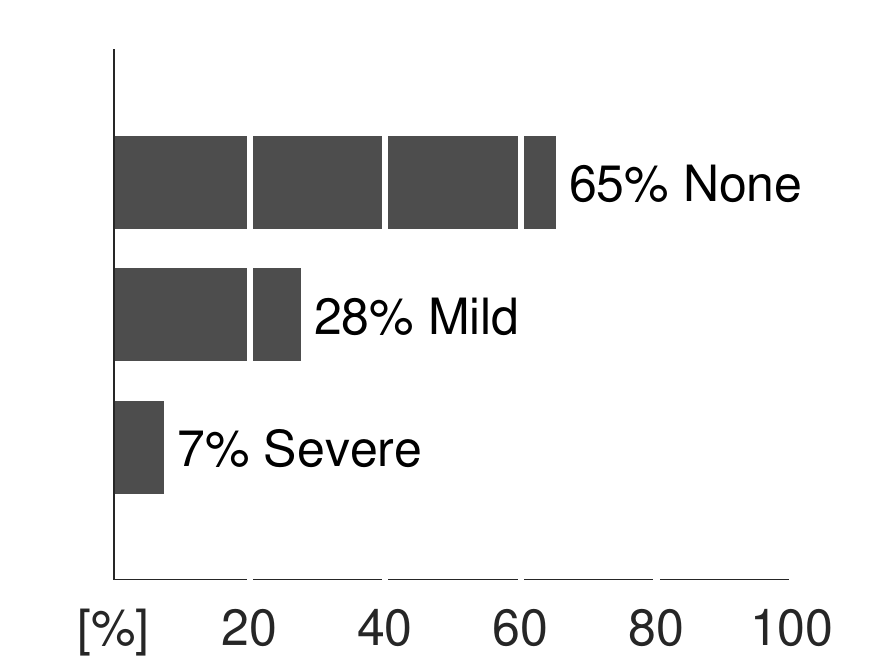}
			\label{img:db_overlap}
		}}%
		\\%
		\subfloat[Head Pitch]{\makebox[0.23\textwidth][c]{
			\includegraphics[width=0.23\textwidth]{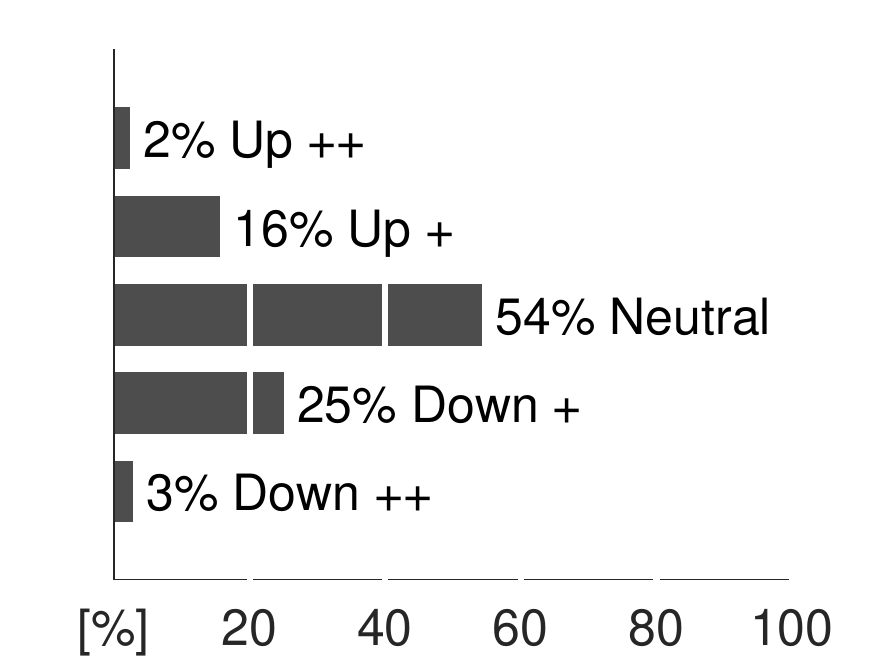}
			\label{img:pitch}
		}}\hfill
		\subfloat[Head Roll]{\makebox[0.23\textwidth][c]{
			\includegraphics[width=0.23\textwidth]{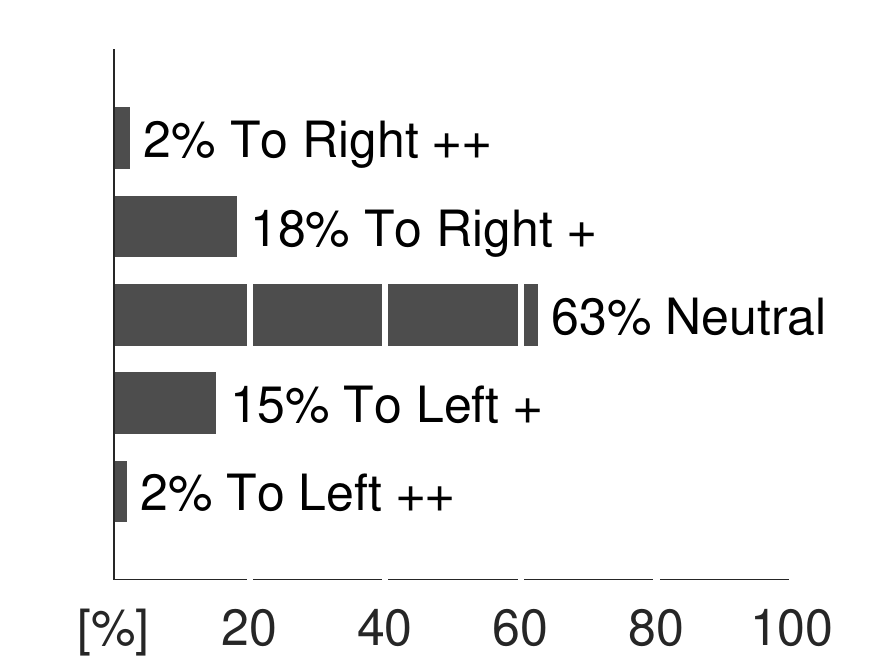}
			\label{img:db_roll}
		}}\hfill
		\subfloat[Head Yaw]{\makebox[0.23\textwidth][c]{
			\includegraphics[width=0.23\textwidth]{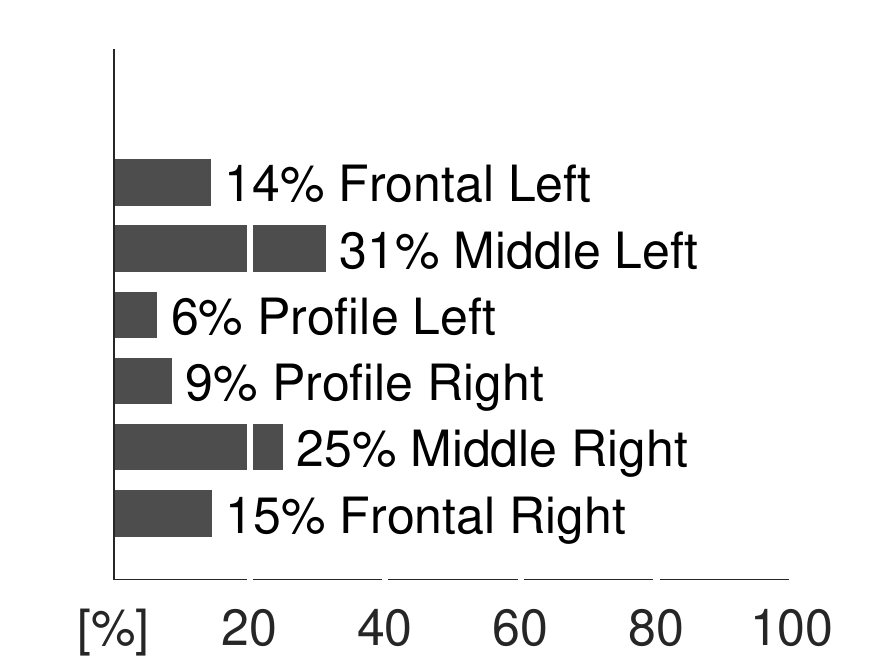}
			\label{img:db_yaw}
		}}\hfill
		\subfloat[Head Side]{\makebox[0.23\textwidth][c]{
			\includegraphics[width=0.23\textwidth]{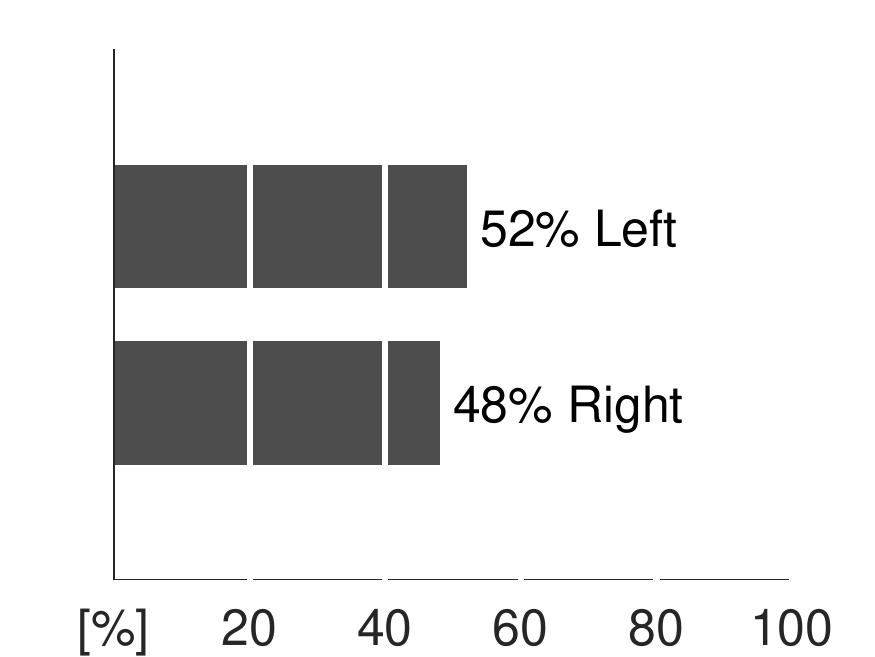}
			\label{img:db_direction}
		}}
		\caption{Distribution of image labels in the Annotated Web Ears (AWE) dataset.}
		\label{img:db_1000}
\end{figure*}

\textbf{Experimental protocols:} We propose two types of experimental protocols for the dataset to enable comparisons of techniques evaluated on our dataset. The first corresponds to protocols for identification (i.e., one-vs-many) experiments and the second to protocols for verification (i.e., one-vs-one) experiments.

For both types of protocols we first partition the data into a \textit{development set} that contains $60\%$ of all images and a \textit{test set} that contains the remaining $40\%$ of images. We propose to use a 5-fold cross validation procedure on the development set to train potential background models, subspaces or classifiers and then apply the trained models on the (hold-out) test set for evaluation. Results
should be reported on the development set by presenting
all performance metrics in the form of means
and standard deviations computed over all 5-folds. Similarly, results on the test set should again be presented in the form of mean values with corresponding standard deviations of all performance metrics, but this time the statistics should be generated through bootstrapping. Files containing lists for the experiments are distributed with our dataset.

The following performance curves and metrics should be generated when reporting performance~\cite{Jain2011}:
\begin{itemize}
  \item \textit{Identification Experiments}: \vspace{-2mm}
  \begin{itemize}
      \item Cumulative Match-score Curves (CMCs),\vspace{-1mm}
      \item Rank-1 recognition (R1) rates,
  \end{itemize}
  \item \textit{Verification Experiments}: \vspace{-2mm}
  \begin{itemize}
      \item Receiver Operating Characteristics (ROC) curves,\vspace{-1mm}
      \item Equal Error Rates (EERs)\del{ and the Verification Rate at 0.1\% False Acceptance Rate (VR-01)}.
  \end{itemize}
\end{itemize}


\subsection{The \dname Toolbox}

\textbf{Motivation:}
Reproducing experimental results is one of the most important steps in many research fields related to computer vision and machine learning. We typically observe a gap between methods described in the literature and available open source implementations~\cite{KlotzOpenBR}, which makes implementing prior work a standard task accompanying every research project. This process is not only time consuming but also affects the reproducibility of published results. To address this problem for the field of ear recognition, we introduce in this section the \dname toolbox, an open source Matlab toolbox designed for reproducible research in ear recognition (and beyond).

\textbf{Software requirements:}
The \dname toolbox is written entirely in Matlab and has been fully tested on the 2015a and 2015b editions, but it should run with older versions as well. To access all of the functionality of the toolbox, Matlab's Computer Vision System Toolbox needs to be installed, as some of the feature extraction techniques supported by the \dname toolbox rely on Matlab's built-in implementations. The toolbox itself exploits a few external libraries but these ship with the toolbox so there is no need for manual installation.


\textbf{Structure and Capabilities:}
The toolbox consists of four major parts devoted to: \textit{i)} data handling,  \textit{ii)} feature extraction, \textit{iii)} distance calculation or/and classification, and \textit{iv)} result visualization. 

The first part of the toolbox (devoted to data handling) takes an arbitrary dataset as an input. The dataset needs to be organized into a directory structure with each folder containing samples of only one class. In this regard, the toolbox is not limited solely to ear images, but can also be used with other biometric modalities. Using functions from the first toolbox part, images are read, normalized and cached for faster reuse at a later time. Currently supported normalization techniques include more than 20 state-of-the-art photometric normalization techniques, which are provided by the  INFace toolbox\footnote{Available from: \url{http://luks.fe.uni-lj.si/sl/osebje/vitomir/face_tools/INFace/}} that is integrated with our toolbox.

The second part of our toolbox is oriented towards feature extraction. For the intended work-flow of the toolbox preselected feature extraction techniques are applied to all normalized images of the selected dataset. The computed feature vectors are (by request) stored in comma-separated-values (CSV) files so that users can either use them in subsequent experiments or export them to other processing or visualization tools. The toolbox currently implements 8 descriptor-based feature extraction techniques, but users can easily plug in their own methods or add functionality to the existing code. The toolbox currently supports extraction of LBP, LPQ, RILPQ, BSIF, DSIFT, POEM, Gabor and HOG features.

The third part of the toolbox, intended for matching and classification of the computed feature vectors, supports two options: \textit{i)} feature vectors can either be matched directly using various distance measures, or \textit{ii)} classifiers may be trained on the training data of the datasets to classify samples. While the toolbox currently supports several distance measures for the first option, we have not included any specific classifiers in the toolbox yet as many ship with every Matlab installation.

The last part of the toolbox generates all relevant performance curves and metrics, such as ROC plots, CMC plots, histograms of client and impostor score distributions, rank-1 recognition rates, equal error rates, and similar metrics useful for comparisons.

The entire processing pipeline implemented in the four parts of the toolbox may be easily configured in any combination and automatically executed for any dataset provided to the toolbox.

One of the biggest advantages of the \dname toolbox is speeding up of the time needed to develop new methods for ear recognition. Several state-of-the-art techniques are already implemented, which makes comparisons straight-forward and the results reproducible. The toolbox is open source so it can be modified, shared and used freely without limitations.



\begin{table*}
\renewcommand{\arraystretch}{2}
\caption{Summary of implemented techniques used for the comparison. The symbol ``pix'' stands for ``pixels''.}
\label{Tab:_summary}
\small
\centering
\begin{tabular}{m{0.22\textwidth}|m{0.7\textwidth}}
\small Method    &   \small Detailed information\\
\hline
LBP~\cite{Pflug2014a,Guo2008,Benzaoui2015,Pietikainen2011}							& uniform LBP, radius: $2$ pix, neighborhood size: $8$, block size: $8 \times 8$ pix, no block overlap\\ \hline
LPQ~\cite{Ojansivu2008}														& window size: $5 \times 5$ pix, block size: $18 \times 18$ pix, no block overlap\\ \hline
BSIF~\cite{Pflug2014a,Kannala2012}												& \#filters: $8$, filter size: $11 \times 11$ block size: $18 \times 18$ pix, no block overlap\\ \hline
POEM~\cite{Vu2010}															& \#orientations: $3$, uniform LBP, radius: $2$ pix, neighborhood size: $8$, block size: $12 \times 12$ pix, no block overlap\\ \hline
HOG~\cite{Pflug2014a,DamerHOG2012,DalaHOG2005}									& cell: $8 \times 8$ pix, block: $16 \times 16$ pix, block overlap: $8$ pix\\ \hline
DSIFT~\cite{Dewi2006,Arbab-Zavar2008,Bustard2010,Krizaj2010,Lowe2004,MoralesSIFT2013}	& \#grid points: $100$, patch size: $16 \times 16$ pix, bin size: $8$\\ \hline
RILPQ~\cite{OjansivuRILPQ2008}													& window radius: $6$ pix, \#angles: $12$, block size: $16 \times 16$ pix, no block overlap\\ \hline
Gabor~\cite{Kumar2007,Kumar2012,Nanni2009,Xiaoyun2009,Meraoumia2015,Daugman1985,StrucGabor2009,Struc2009,Struc2010}    & \pbox{13cm}{\#orientations: $8$, \#scales: $5$, \#filters $40$, down-sampling factor: $64$}
\end{tabular}
\end{table*}

\section{Experiments and Results}
\label{section:experiments}


To highlight the current capabilities of the \dname toolbox and provide an independent comparative analysis of several state-of-the-art descriptor-based ear recognition techniques we conduct in this section recognition experiments with \del{four}\add{two} popular datasets as well as our newly introduced \dname dataset. All presented experiments are fully reproducible using the \dname toolbox and rely on the same experimental protocol.

%
%

\subsection{Comparative Assessment on Existing Datasets}
Many existing ear datasets do not define an experimental protocol that can be used during experimentation. Consequently, authors set up their own protocols, which makes comparisons of published results difficult. To establish an objective ranking of the state-of-the-art (descriptor-based) methods for ear recognition, we conduct verification as well as identification experiments with 8 different techniques from the \dname toolbox on \del{four}\add{two} popular ear datasets. We follow a similar processing pipeline for all techniques, which: \textit{i)} rescales the images to a fixed size of $100\times100$ pixels, \textit{ii)} corrects for illumination-induced variability by applying histogram equalization to the resized images, \textit{iii)} subjects the images to the selected feature (or descriptor) extraction techniques, and \textit{iv)} produces a similarity score for each probe-to-gallery comparison by computing the distance between the corresponding feature vectors.

A brief summary of the feature extraction techniques and selected hyper-parameters used in the experiments is given in Table~\ref{Tab:_summary}. We implement feature extraction techniques based on LBP, LPQ, BSIF, POEM, HOG, DSIFT, RILPQ and Gabor features and select the open hyper-parameters through cross-validation. We use the chi-square distance to measure the similarity between the feature vectors of the probe and gallery images for all histogram-based descriptors and the cosine similarity measure for the Gabor features.

\begin{figure}[!htb]
\captionsetup{type=figure}
	\center
        \subfloat[IITD II CMCs]{\makebox[0.5\columnwidth][c]{
		\includegraphics[width=0.5\columnwidth,trim=0.2cm 6cm 2cm 7cm,clip]{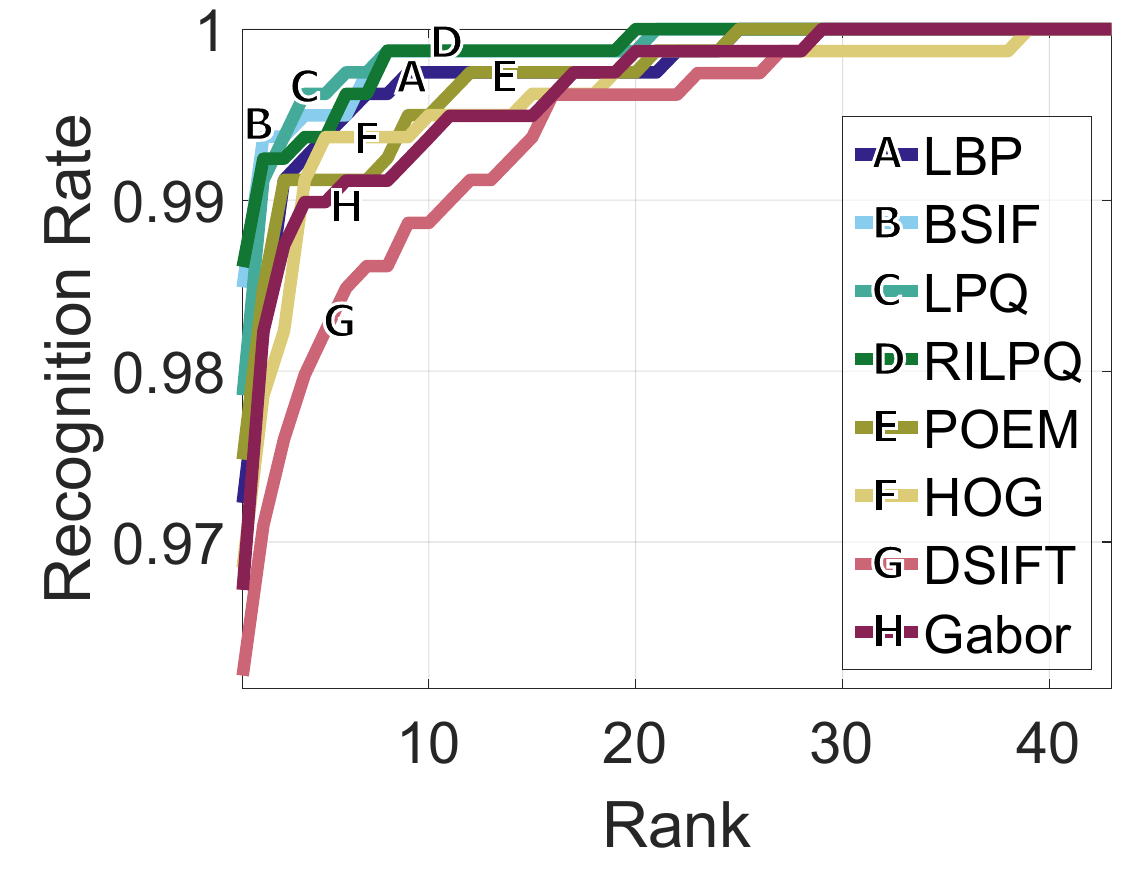}
		\label{img:results_p1_3}
	}}
        \hfil
        \subfloat[USTB II CMCs]{\makebox[0.5\columnwidth][c]{
		\includegraphics[width=0.5\columnwidth,trim=0.2cm 6cm 2cm 7cm,clip]{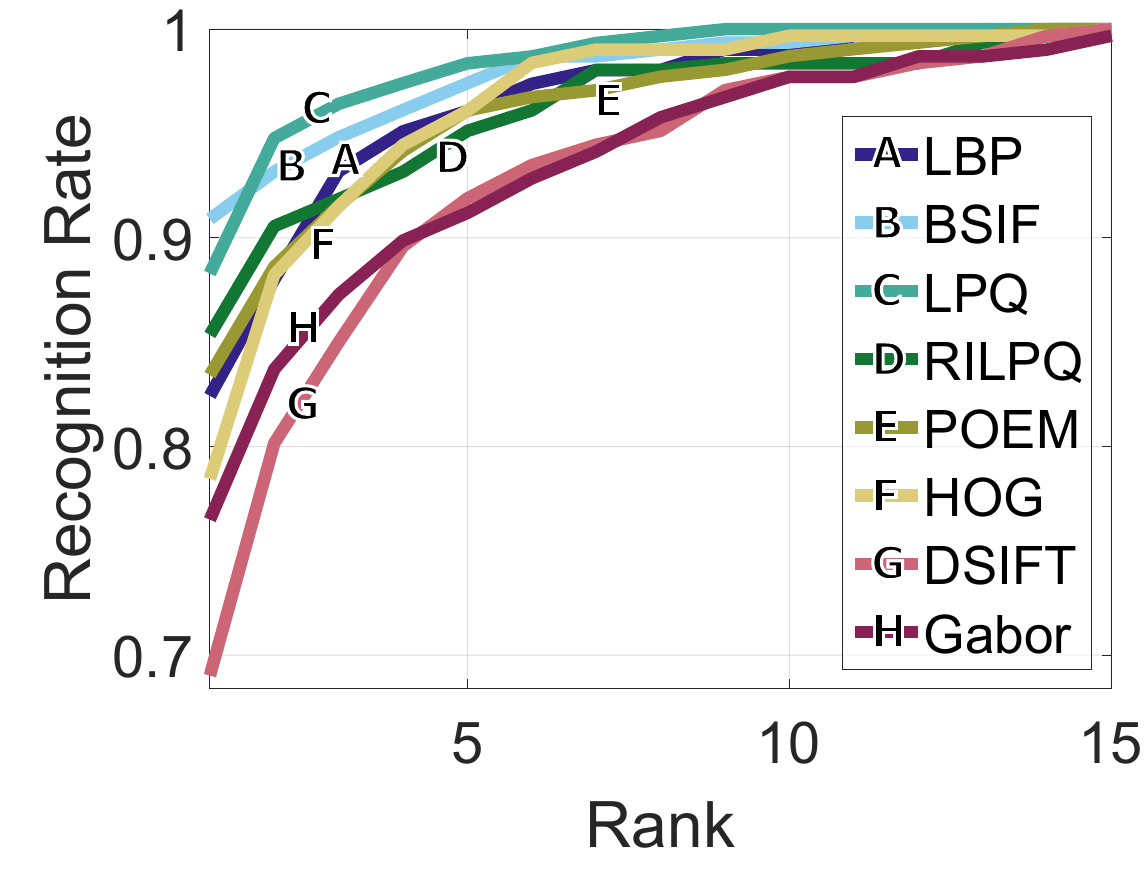}
		\label{img:results_p1_7}
	}}
	\\
	\subfloat[IITD II ROC curves]{\makebox[0.5\columnwidth][c]{
		\includegraphics[width=0.5\columnwidth,trim=0.2cm 6cm 2cm 7cm,clip]{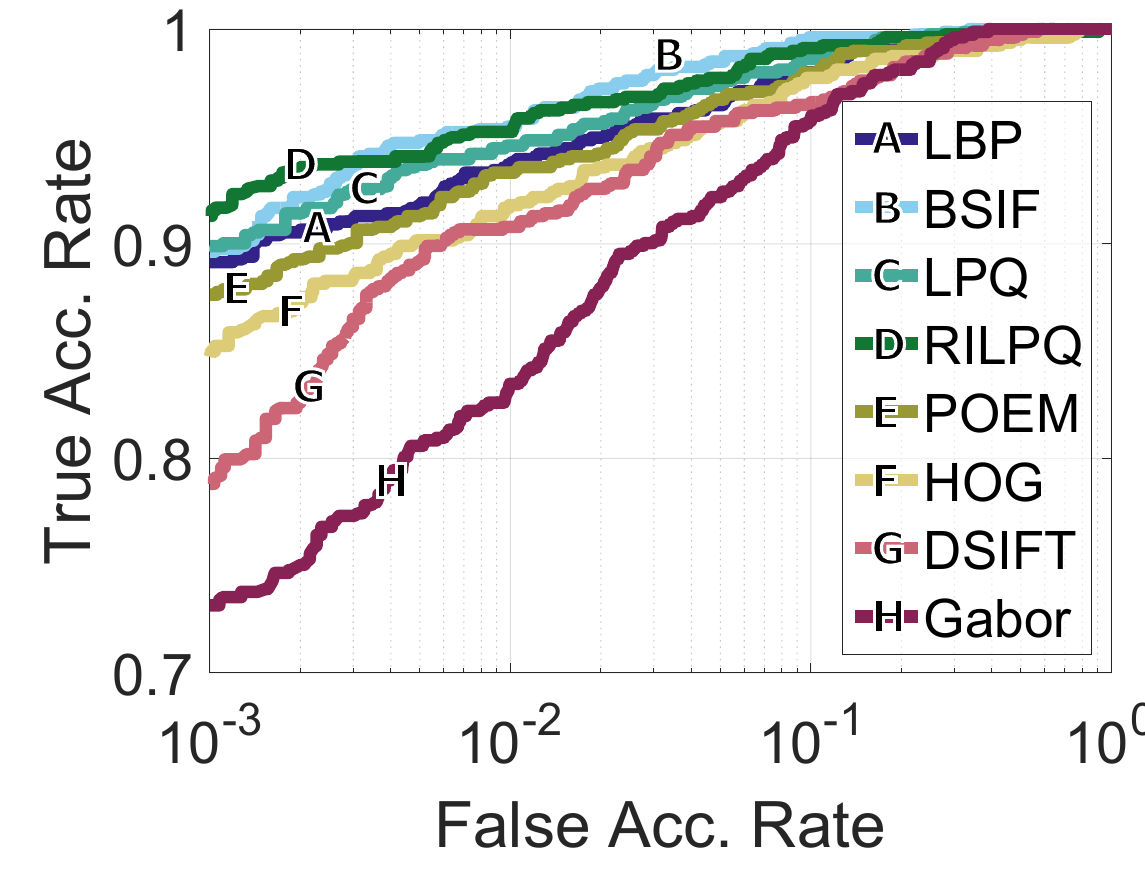}
		\label{img:results_p1_4}
	}}			
	\hfil
	\subfloat[USTB II ROC curves]{\makebox[0.5\columnwidth][c]{
		\includegraphics[width=0.5\columnwidth,trim=0.2cm 6cm 2cm 7cm,clip]{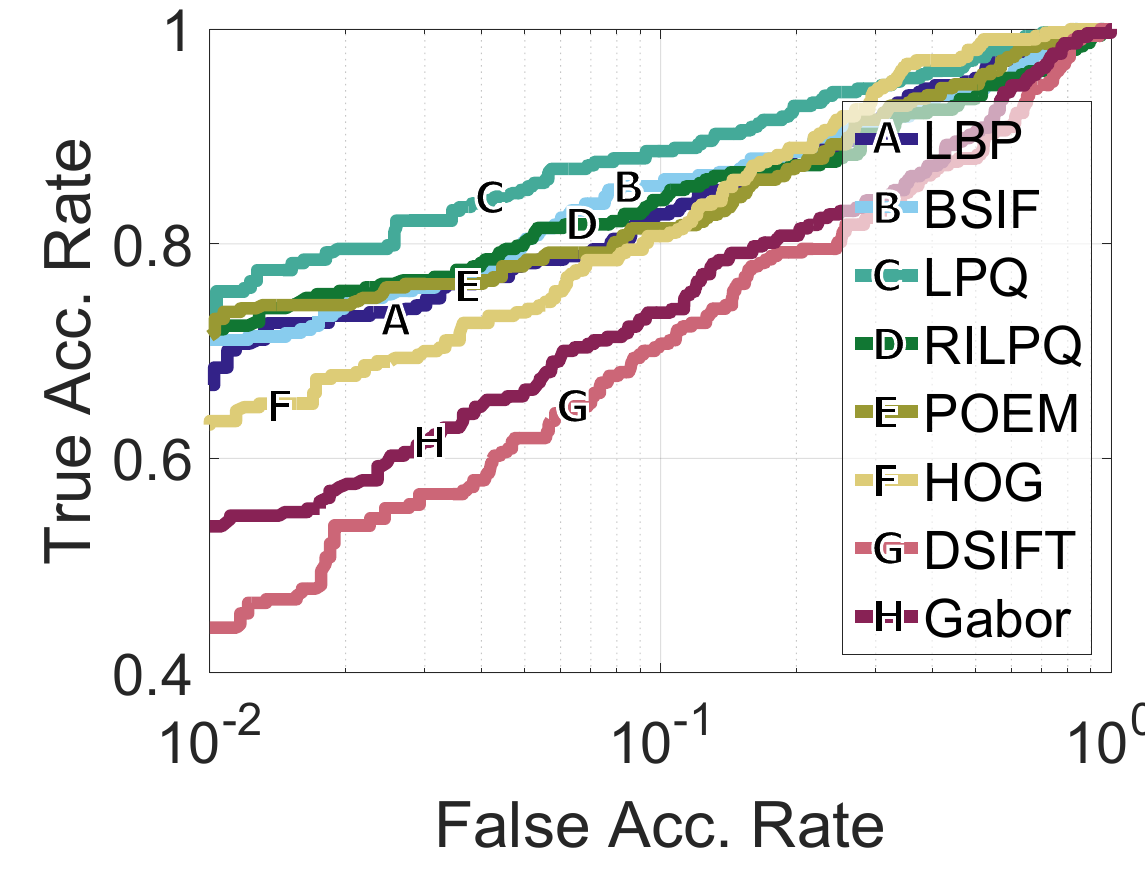}
		\label{fig:results_p1_8}
	}}
	\caption{Comparative assessment of 8 state-of-the-art techniques. The upper row shows CMC curves and the lower row ROC plots for the two datasets. Experiments on the IITD II dataset were conducted with pre-aligned images, while the results on the USTB II datasets were generated with non-aligned images. Note that the performance of all assessed techniques is better on the aligned than on the nonaligned dataset and not much room is left for further improvements.}
	\label{fig:results_p1}
\end{figure}

We perform the comparative assessment on images from the IITD II and USTB II datasets\footnote{We also performed experiments on the IITD I and USTB I datasets with comparable results but, for the sake of brevity, do not report the results here.}. These datasets are among the most popular datasets for ear recognition and have, therefore, also been selected for our experiments. The reader is referred to Section~\ref{section:data} for a more in-depth description of these datasets.

Two types of experiments are conducted. The first uses the cropped and pre-aligned images that ship with the IITD II dataset, while the second employs cropped but non-aligned images from the USTB II dataset\del{s}. As already suggested above, the goal of these experiments is to study the behavior of the state-of-the-art descriptor-based methods on images of varying quality and to demonstrate the current capabilities of our \dname toolbox.

We conduct 5-fold cross validation experiments on \del{all four}\add{the two} datasets and generate average ROC and CMC plots over all 5 folds as shown in Fig.~\ref{fig:results_p1}. We adjust the scale on the $y$-axis for each dataset to better highlight the differences in the performance of the assessed techniques. To avoid clutter in the graphs, we do not provide error bars for the performance curves but instead include standard deviations for the performance metrics in \del{Tables~\ref{Tab:_P1aligned} and~\ref{Tab:_P1nonaligned}} \add{Table~\ref{Tab:_P1}}. Here, we report the mean rank-1 recognition rate (R1) for the identification experiments and the equal error rate (EER) for the verification experiments.\del{ and the verification rate at 0.1\% false acceptance rate (VR-01)  The USTB datasets do not contain enough samples to report VR-01 rates, so we report the verification rates at 1\% false acceptance rate (VR-1) instead}.

\begin{table}[!htb]
\renewcommand{\arraystretch}{1.1}
\caption{Comparative assessment of 8 state-of-the-art techniques on the pre-aligned images from the IITD II dataset and the non-aligned images from the USTB II dataset. For each performance metric the results are provided in the form of the mean and the standard deviation over 5-folds. The performance of the assessed techniques on both datasets is comparable to the existing state-of-the-art from the literature.}
\label{Tab:_P1}
\centering
\footnotesize
\begin{tabular}{| l | cc | cc |}
\hline
\multirow{ 3}{*}{\small Method}    &   \multicolumn{2}{c|}{\small IITD II} &   \multicolumn{2}{c|}{\small USTB II}\\
\cline{2-5}
                & \small R1 & \small EER & \small R1 & \small EER \\
\hline	
\small LBP		& $97.2\pm0.9$ & $3.9	\pm1.7$ & $82.4\pm7.5$ & $15.0\pm7.5$ \\
\small BSIF		& $98.5\pm1.0$ & $2.4	\pm1.4$ & $90.9\pm6.5$ & $13.4\pm7.2$ \\
\small LPQ		& $97.9\pm0.5$ & $3.3	\pm1.5$ & $88.3\pm7.6$ & $11.1\pm5.6$ \\
\small RILPQ	& $98.6\pm0.9$ & $3.1	\pm1.3$ & $85.4\pm8.7$ & $13.7\pm8.0$ \\
\small POEM	& $97.5\pm1.1$ & $3.9	\pm1.4$ & $83.4\pm7.2$ & $15.3\pm8.3$ \\
\small HOG		& $96.9\pm0.7$ & $4.5	\pm1.7$ & $78.5\pm7.0$ & $14.6\pm7.2$ \\
\small DSIFT	& $96.2\pm1.9$ & $4.5	\pm2.0$ & $69.0\pm9.1$ & $20.5\pm8.5$ \\
\small Gabor	& $96.7\pm1.6$ & $6.6	\pm1.6$ & $76.5\pm4.4$ & $19.3\pm7.6$ \\
\hline
\end{tabular}
\end{table}

The results show that the performance of all assessed techniques is significantly better for aligned than for non-aligned images. This is expected as pose variability is generally considered one of the major factors influencing ear recognition performance. 

On the aligned images all evaluated techniques perform similarly in the identification experiments $\bigl($see Fig.~\ref{fig:results_p1} -- (a)\del{, (b)}$\bigr)$. The difference in the recognition performance is in fact minimal for techniques that rely on any of the binary pattern variants to encode the ear texture. For the verification experiments $\bigl($Fig.~\ref{fig:results_p1} -- \del{(e), (f)}\add{(c)}$\bigr)$ the differences among the techniques are somehow larger. Here,  HOG, DSIFT and Gabor features result in lower performance, while the binary-pattern-based techniques again perform better and more or less on the same level.

On the non-aligned images the ranking of the techniques does not change significantly compared to the aligned images. Techniques based on binary patterns perform very similarly, 
while HOG, DSIFT and Gabor-based techniques again result in worse performance $\bigl($Fig.~\ref{fig:results_p1} -- \del{(c), (d), (g), (h)}\add{(b), (d)}$\bigr)$. The main conclusion that we can draw from this series of experiments is that techniques based on binary patterns should be preferred over other local methods, while the choice of a specific method within this group should be made based on other non-performance related criteria.

The presented results are comparable to what was reported by other authors for similar techniques on the same datasets, e.g.,~\cite{Benzaoui2014,Pflug2014a,Guo2008,Benzaoui2015a,Meraoumia2015}, which suggests that the  \dname toolbox provides competitive implementations of all evaluated descriptor-based techniques. 

%
%
%

\begin{figure}[!htb]
\captionsetup{type=figure}
	\center
		\subfloat[\dname CMCs]{\makebox[0.5\columnwidth][c]{
			\includegraphics[width=0.5\columnwidth,trim=0.2cm 6cm 2cm 7cm,clip]{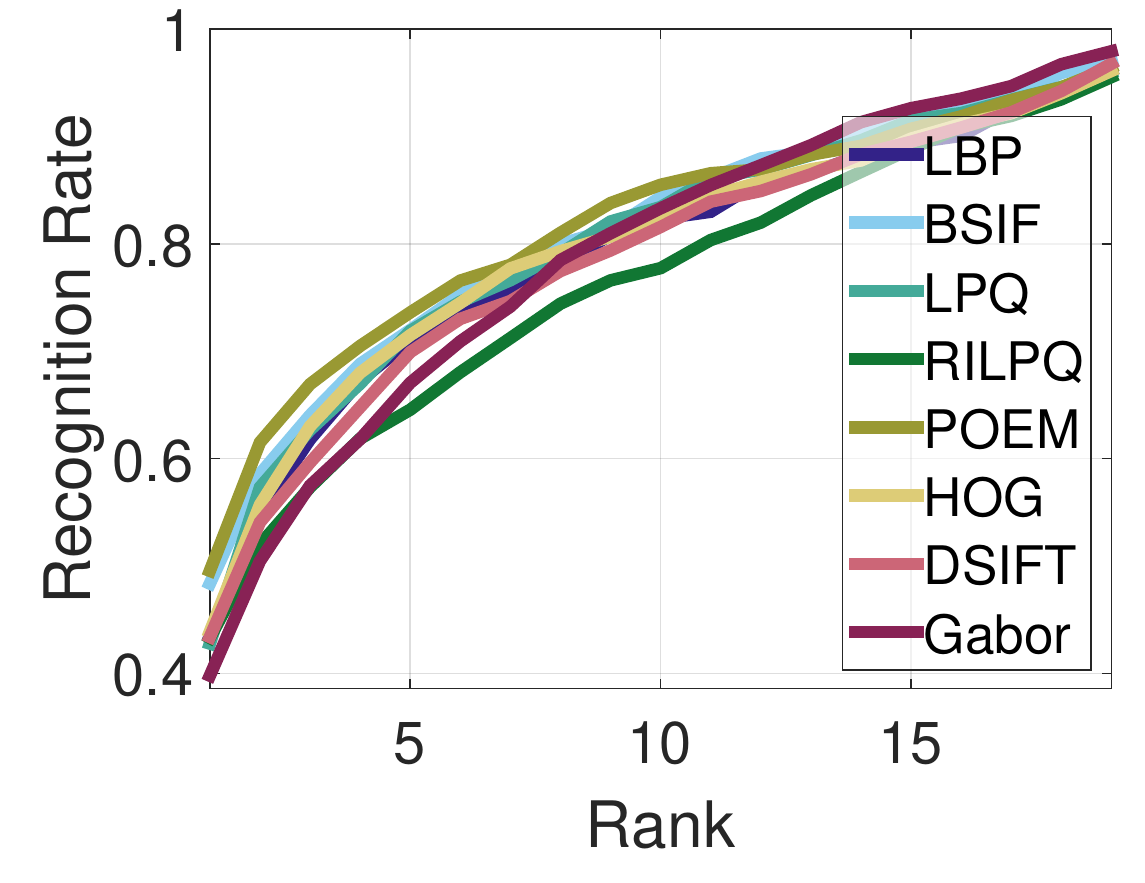}
			\label{img:impostor1}
		}}
		\hfil
		\subfloat[\dname ROC curves]{\makebox[0.5\columnwidth][c]{
			\includegraphics[width=0.5\columnwidth,trim=0.2cm 6cm 2cm 7cm,clip]{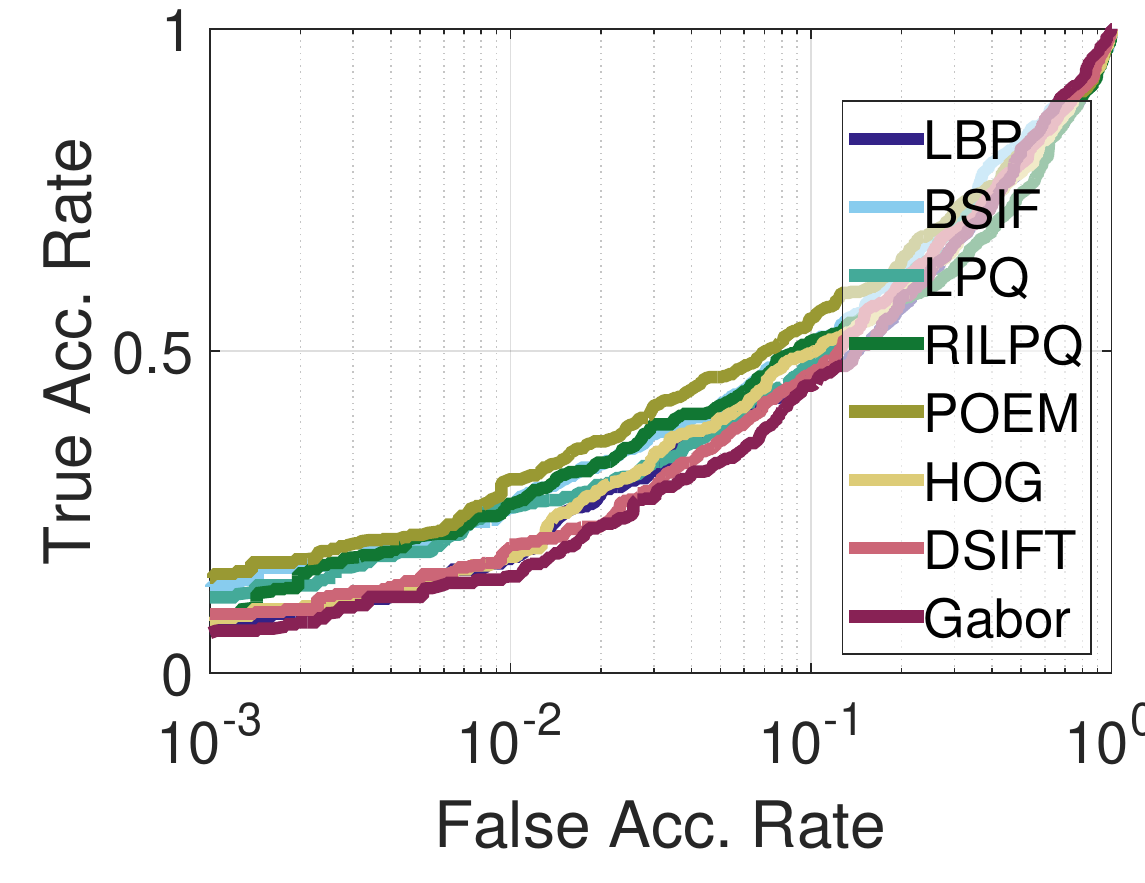}
			\label{img:impostor2}
		}}
		\caption{Performance curves generated during 5-fold cross-validation experiments on the development part of the Annotated Web Ears (AWE) dataset. The left graph depicts CMC and the right graph  ROC curves of the experiments. The results show the usefulness and the difficulty of the AWE dataset and indicate that there is significant room for improvement of ear recognition technology.  Here, the performance curves are not marked as in Fig. 7 due to the almost equal performance of all descriptors.}
		\label{img:impostors}
\end{figure}

\subsection{Comparative Assessment on the \dname Dataset}

We repeat the experiments from the previous section with the same 8 feature extraction techniques, i.e., based on LBP, LPQ, RILPQ, BSIF, DSIFT, POEM, Gabor, and HOG features, on the \dname dataset and follow the experimental protocols outlined in Section~\ref{Section:DatasetAndToolbox}. We use the same hyper-parameters as in the previous section for all techniques.  The goal of these experiments is to demonstrate the difficulty of recognizing ear images captured in completely uncontrolled conditions and to establish the performance of the state-of-the-art descriptor-based methods on images captured in the wild. The images of the \dname dataset are not aligned; the results presented in this section should, therefore, be compared with the results reported on the \del{IITD I and} USTB II dataset\del{s} when assessing the difficulty of the new dataset.

Following the predefined experimental protocols, we split the data into two parts and \del{first} conduct experiments on the development part of the dataset using a 5-fold cross-validation procedure. As suggested in Section~\ref{Section:DatasetAndToolbox}, files containing lists of the required experiments for each fold are distributed with the dataset. We report the results in the form of CMC and ROC curves and with the predefined performance metrics. When reporting quantitative results, we provide mean and standard deviations for each required performance metric computed over all 5-folds. We conduct a total of $600$ identification experiments and $1,856$ legitimate vs. $34,308$ illegitimate authentication trials for the verification experiments.

\add{Note that the dataset partitioning into separate development and test sets is meant to facilitate training of classification techniques (such as~\cite{neurocomputing1}). In this case the development set is intended for model learning and parameter tuning, while the test set is reserved for the final evaluation. However, because we do not use classification models in our experiments, we do not report the results on the test set in this paper, but only state that they are similar to the results obtained on the development set.}

For the development set the results are presented in Fig.~\ref{img:impostors} and Table~\ref{Tab:_P2}. As can be seen, the performance of all assessed techniques on the \dname dataset is significantly lower than on the datasets used in the previous section for both the identification and verification experiments. The mean rank-1 recognition rates range from 39.8\% achieved with Gabor features to 49.6\% achieved by the POEM descriptor. The remaining recognition rates are close and all above 40\%. For the verification experiments the performance of all evaluated methods is also close but, considering the entire ROC curve, the POEM descriptor is again the top performer. Here, the best performance with respect to EER is around 30\%. These results highlight the difficulty of the \dname dataset and clearly show that more research is needed to make the ear-recognition technology applicable to completely uncontrolled images. 


\begin{table}[!htb]
\renewcommand{\arraystretch}{1.1}
\caption{Comparative assessment of 8 state-of-the-art techniques on the development part of the Annotated Web Ears (AWE) dataset. For each performance metric the results are provided in the form of the mean and the standard deviation over 5-folds generated during cross-validation experiments. The results demonstrate the usefulness and the difficulty of the
dataset.}
\label{Tab:_P2}
\centering
\footnotesize
\begin{tabular}{ l|  cc}
\small Method & \small R1 & \small EER  \\
\hline
LBP			& $43.5\pm7.1		$ & $32.0\pm7.4		$   \\
BSIF		& $48.4\pm6.8		$ & $30.0\pm9.6		$   \\
LPQ			& $42.8\pm7.1		$ & $30.0\pm7.4		$   \\
RILPQ		& $43.3\pm9.4		$ & $34.0\pm6.4		$   \\
POEM		& $49.6\pm6.8		$ & $29.1\pm9.1		$   \\
HOG			& $43.9\pm7.9		$ & $31.9\pm7.8		$   \\
DSIFT		& $43.4\pm8.6		$ & $30.9\pm8.4		$   \\
Gabor		& $39.8\pm7.1		$ & $32.1\pm8.1		$   \\
\end{tabular}
\end{table}

\del{This finding is further supported by the experiments on the test part of the \dname dataset. The results for this part are again generated in accordance with the predefined experimental protocols. To produce confidence estimates for each performance metric, we sub-sample the similarity scores generated during the experiments using a Monte-Carlo sub-sampling procedure. We generate 100 bootstrap sets, each containing 60\% of all similarity scores. From these bootstrap sets, we compute the mean performance curves and metrics with corresponding standard deviations that are shown in Fig.~\ref{img:impostorsX} and Table~\ref{Tab:_P3}.}


\del{As can be seen, the performance on the test part of the dataset is even lower than on the training part. The confidence bounds here are also tighter, which is a consequence of the bootstrapping procedure that is known to produce conservative confidence estimates. Here, the best identification performance in terms of the rank-1 recognition rate is again achieved by the POEM descriptor with 19.7\%. However, the difference in the recognition performance between all assessed techniques is neglectable and is not significant from a statistical, let alone practical point of view. Similar observation could also be made for the verification experiments.

These results are in line with the findings on the development set and again indicate that ear recognition has still a long way to go before it can be used in a fully automatic manner with real world imagery. We believe that the \dname dataset is an important tool for this purpose and hope that it will help the community to advance the technology further.

In the next section we try to identify some of the most important open problems in the field of ear recognition and outline promising research directions that we feel could help to improve the existing technology.}


\section{Open Questions and Research Directions}
\label{section:research}


Ear recognition still represents a relatively young research area compared to other biometric modalities. Problems and research questions that are well studied in other fields require more research and provide room for exploration~\cite{neurocomputing3,neurocomputing4,neurocomputing5}. In this section we briefly describe what we feel are the most important open problems and most promising research directions in the field. Our goal here is to establish a list of topics worth investigating in the years to come.

\subsection{Ear Detection and Alignment}

Fully automatic ear recognition systems require reliable ear detection techniques capable of determining the location of the ear in the input images (or video) in an efficient and possibly real-time manner. While there has been some research effort in this direction (see, for example,~\cite{NixonSurvey2013,Pflug2012} for recent surveys related to detection), ear detection is still an unsolved problem and no widely adopted solution has been proposed yet in the literature. In our opinion, two major issues need to be addressed to foster advancements in this area. The first is the introduction of large-scale ear datasets captured outside laboratory environments that would allow training competitive ear detectors to be capable of operating on real-world imagery and video footage. As a shortcut towards addressing this issue, the existing face datasets with suitable profile images could be used for this purpose. The second issue is the implementation of publicly available open-source tools for ear detection that would provide baselines for experimentation and development. To the best of our knowledge no competitive open-source implementation of ear detection techniques can currently be found on-line.

Another major problem in the automatic ear recognition systems is alignment. Pose variations are generally considered to be one of the biggest issues plaguing the technology~\cite{NixonSurvey2013,Pflug2012}. In-plane rotations as well as moderate out of plane variations can effectively be addressed with alignment techniques, which makes alignment a key factor of robustness for ear recognition. Manual ear alignment is common, but existing automatic alignment techniques include active shape models~\cite{Yuan2007} or combinations of active contours models and ovoids over ear contours~\cite{Gonzalez2012Normalization}. However, recent advancements in boosted/cascaded landmark detection and alignment, e.g.,~\cite{XiongD13,Cihan2015}, can also be used for ear images provided that appropriately annotated data for training becomes available. Some initial attempts in this direction have been presented in~\cite{Pflug2014a}.

\subsection{Handling Ear Variability}

Appearance variability encountered with ear images is induced by different factors, such as occlusions by hair, clothing or accessories, self-occlusions due to view-point changes, or pose variations (among others). These factors represent the main sources of errors in ear recognition systems and therefore need to be addressed appropriately.

Various techniques have been proposed in the literature capable of handling moderate occlusions (by external objects or due to view point changes), e.g.,~\cite{Arbab-Zavar2008,Bustard2010}, though it is not clear where the breaking points of these techniques are or how much occlusion can be tolerated by the existing technology. Ear occlusion does not only affect recognition schemes but likely has an equally large effect on ear detection and alignment techniques. Errors from these steps are propagated through the processing pipelines and degrade the recognition performance of automatic ear recognition systems. Potential solutions here could exploit contextual information and address occlusions by modeling the entire facial profile instead of the ear region alone. A similar idea was, for example, successfully exploited in~\cite{BustardNixon2010}.

Pose variations can to a large extent be addressed  with suitable alignment techniques, as already emphasized in the previous section. 3D data and generic 3D models that could be exploited to normalize the geometry of the ear may play a bigger role in the future. Especially appealing in this context are techniques based on the recently popularized concept of analysis-by-synthesis~\cite{Tenenbaum2015} where images are not aligned into predefined canonical forms. Instead the poses of probe and gallery images are matched through a reposing procedure that renders one of the images in the pose of the second. This pair-wise approach may represent a viable research direction for pose-invariant ear recognition in the future.

\subsection{Feature Learning}

Ear recognition approaches mainly followed the development of other computer vision fields and progressed from early geometric techniques and holistic methods to more recent local and hybrid approaches. The next step in this development is the use of feature learning techniques based on convolutional neural networks (CNNs) for ear image description or even end-to-end system design. While techniques based on neural networks are becoming increasingly popular in other areas of computer vision and machine learning, ear recognition has not yet benefited from recent advances in this area mainly due to the lack of large-scale datasets. 

\del{Only a handful of methods in the literature use deep neural networks for ear recognition. This fact can mainly be attributed to the lack of large-scale datasets (with hundreds of thousands of ear images) that are needed to train competitive deep models. We believe that in the short-to-medium term domain adaptation techniques will be adopted to make use of available deep models, while in the long term, larger datasets will have to be introduced to benefit from advances in deep learning.}


\subsection{Large-scale Deployment}

To date, all available datasets for evaluating ear recognition technology are still relatively modest in size, featuring images of at most a few hundred subjects. It is still not clear how the existing technology scales with the number of subjects, especially for large-scale identification tasks. Experimental studies on this topic are needed as well as mathematical models of ear individuality that would lay the theoretical basis for the distinctiveness of ears~\cite{NixonSurvey2013}. Some research on this topic can be found in the forensic literature but more work is required to provide empirical evidence for the suitability of ear images for large-scale identification tasks.

\subsection{Imaging Modalities}

Most research work related to ear recognition focuses on 2D images captured with common off-the-shelf (COTS) cameras or CCTV systems. However, other modalities, such as 3D ear data or ear-prints, are equally important and offer similar research challenges as 2D images. Detection, segmentation, feature extraction and modeling techniques need to be developed for these modalities and we believe that research focused on these problems will become more important in the future.

Another interesting problem that we believe will garner more interest in the long term is heterogeneous ear recognition where ear images captured with different imaging modalities will be matched against each other.

\subsection{Understanding Ear Recognition}

As pointed out by~\cite{NixonSurvey2013,Pflug2012}, there are several factors pertaining to ear recognition that are not yet well understood  and offer room for further exploration. These factors include: \textit{i) Ear symmetry:} some studies have found that the left and right ear are close to symmetric for many subjects but more research is needed to find ways of exploiting this fact in automatic recognition systems; \textit{ii) Aging:} it is known that the size of the ear increases with age, e.g.,~\cite{Sforza2009,Gualdi1998,Ferrario1999}, but how this effect influences the recognition performance of automatic techniques is still unclear; the main problem here is the lack of suitable datasets that would be needed for a detailed analysis of the effects of template aging; \textit{iii) Inheritance:} as pointed out by~\cite{Pflug2012}, the work of Iannarelly~\cite{Iannarelli1989} suggested that some characteristics of the outer ear can be inherited; whether these characteristics could be identified by automatic techniques and exploited to determine if two persons are related or whether they could present a source of error for automatic recognition systems is still an open question.

\section{Conclusion}
\label{section:conclusion}

\del{In this paper we surveyed the field of ear recognition. We discussed the early developments pertaining to ear recognition and presented a taxonomy and high-level comparison of existing techniques. We reviewed the most popular datasets used in ear recognition research and introduced a new, publicly available ear dataset gathered from the web. The dataset is the first of its kind and represent a considerable challenge for the existing technology. We demonstrated the difficulty of the dataset in recognition experiments with a toolbox of state-of-the-art ear recognition techniques that was also introduced in this paper. Finally, we identified several open problems in the field and provided an outlook into the future of ear recognition.}
\add{In this paper we surveyed the field of ear recognition including the most popular datasets used in the ear recognition research, and introduced a new, publicly available ear dataset gathered from the web. The dataset is the first of its kind and represents a considerable challenge for the existing technology. We demonstrated the difficulty of the dataset in recognition experiments with a toolbox of the state-of-the-art ear recognition techniques that was also introduced in this paper. Finally, we identified several open problems in the field and provided an outlook into the future of ear recognition.}

We hope that the survey, toolbox and dataset will help the research community to improve the state-of-technology in ear recognition and inspire new research in this area. The toolbox is open to the wider biometric community and useful beyond ear recognition\del{. Our plan for the future is to include more techniques in the toolbox making it even more appealing for research purposes. Among these techniques we plan to focus on} \add{and we plan to include more techniques in the future, such as:} ear detection and alignment approaches, normalization techniques, new descriptors and more complex procedures for feature learning.









\FloatBarrier

\bibliographystyle{elsarticle-num}

\end{document}